\documentclass[10pt,twocolumn,a4paper]{article}

\usepackage{cvpr}

\usepackage{amsmath}
\usepackage{amssymb}
\usepackage{color}
\usepackage{enumitem}
\usepackage{epsfig}
\usepackage{graphicx}
\usepackage{multirow}
\usepackage{placeins}
\usepackage{times}

\definecolor{teal}{rgb}{0,0.5,0.5}
\definecolor{darkgreen}{rgb}{0,0.5,0}
\definecolor{darkblue}{rgb}{0,0,0.7}
\definecolor{revmwcol}{rgb}{0.4,0.5,0.1}
\definecolor{darkred}{rgb}{0.75,0,0}

\newcommand{\Figure}[1]{Figure~\ref{#1}}

\newcommand{\Equation}[1]{Eq.~\ref{#1}}

\usepackage[pagebackref=true,breaklinks=true,colorlinks,bookmarks=false]{hyperref}

\cvprfinalcopy %
\ifcvprfinal\fi

\begin{document}
\title{Deep Non-Line-of-Sight Reconstruction}

\author{Javier Grau Chopite\\
University of Bonn\\
{\tt\small jgraucho@uni-bonn.de}
\and
Matthias B. Hullin\\
University of Bonn\\
{\tt\small hullin@cs.uni-bonn.de}
\and
Michael Wand\\
University of Mainz\\
{\tt\small wandm@uni-mainz.de}
\and
Julian Iseringhausen\\
University of Bonn\\
{\tt\small iseringhausen@cs.uni-bonn.de}
}

\maketitle
\begin{abstract}
The recent years have seen a surge of interest in methods for imaging beyond the direct line of sight. The most prominent techniques rely on time-resolved optical impulse responses, obtained by illuminating a diffuse wall with an ultrashort light pulse and observing multi-bounce indirect reflections with an ultrafast time-resolved imager. 
   {Reconstruction of geometry from such data, however, is a complex non-linear inverse problem that comes with substantial computational demands.   	
   	In this paper, we employ convolutional feed-forward networks for solving the reconstruction problem efficiently while maintaining good reconstruction quality.}
   {Specifically, we devise a tailored autoencoder architecture, trained end-to-end, that maps transient images directly to a depth map representation. Training is done using an efficient transient renderer for diffuse three-bounce indirect light transport that enables the quick generation of large amounts of training data for the network.}
   We {examine the performance of our} method on a variety of synthetic and experimental datasets and {its dependency on the choice of training data and augmentation strategies, as well as architectural features.}
	{We demonstrate that our feed-forward network, even though it is trained solely on synthetic data, generalizes to measured data from SPAD sensors and is able to obtain results that are competitive with model-based reconstruction methods.}
\end{abstract}

\section{Introduction}
Over the last decades, the reconstruction of object shapes by means of time-of-flight measurements has matured to a highly pervasive and impactful technology.
While it can already be challenging to image objects that are directly visible, recent years have seen researchers successfully demonstrating the even harder task of reconstructing targets hidden beyond the direct line of sight. In this setting, which we call the non-line-of-sight (NLoS) reconstruction problem, the loss of temporal and directional information caused by multiple bounces across more or less diffuse surfaces means that the problem has to be approached in a drastically different manner. Among the techniques proposed in literature for reconstructing object geometry and reflectance from three-bounce indirect measurements are combinatorial approaches \cite{Kirmani:2009}, parametric model fits \cite{iseringhausen2018,Kirmani:2009,klein2016tracking,Naik:2011,pediredla2017reconstructing,tsai2019beyond}, backprojection \cite{ArellanoOpEx2017,Laurenzis:2014,Velten:2012:Recovering}, inverse filtering approaches \cite{Heide:2014,otoole2018} and space-time diffraction integrals \cite{Lindell:2019:Wave,liu2019non}. For the purpose of this paper, we focus on such active-illumination, time-of-flight input despite the fact that steady-state or even purely passive approaches have also been demonstrated \cite{bouman2017turning,chen_2019_NLoS,katz2012looking,klein2016tracking}. %

Representation learning methods based on deep neural networks have become a powerful and widely adopted tool for representing complex mappings between high-dimensional (function) spaces of natural signals by learning from examples. In recent work, this includes convolutional regressors for direct time-of-flight imaging \cite{Lindell:2018:3D}. 
Using scale-space representations with short-cuts across matching scales has turned out to be a particularly successful recipe \cite{ronneberger2015unet}. Generative convolutional feed-forward networks have also been used for the creation of 3D data, including 2D-to-3D translations \cite{Dosovitskiy2017,Lin2018learning,Wu2015CVPR}.
Despite this widespread success, no such approaches have so far been applied to the NLoS reconstruction problem. We assume that this can be attributed to the practical difficulty of generating large amounts of NLoS training data in laboratories or natural scenarios. A key step is therefore the development of a simulator that strikes a good balance between computational cost and physical predictiveness so as to keep the training problem tractable while yielding generalizable results. %

In this paper, we build upon a highly efficient transient renderer \cite{iseringhausen2018} to generate large amounts of training data for solving the NLoS reconstruction problem using deep neural networks. Our contributions to arrive at a full deep learning solution include:
\begin{itemize}[leftmargin=*]\addtolength{\itemsep}{-2.5mm}
	\item We propose a parameter-free end-to-end regressor that computes depth maps from 3D space-time impulse responses. 
	\item To train our network, we propose sampling strategies to generate a wide gamut of representative NLoS datasets.
	\item We propose a fast approximate model for real-world time-of-flight setups which is used for data augmentation. The model, while not as refined as existing SPAD models \cite{hernandez2017computational}, can is efficient enough to be applied as an augmentation step during training.%
	\item We evaluate the performance of our system on synthetic as well as experimental input data.
\end{itemize}
We show that our system, while never trained on real-world input, can make meaningful predictions on experimental data. This holds, up to a certain degree, even when scenes contain retroreflective objects that violate the assumptions of our (purely diffuse) forward model. To our knowledge, this constitutes the first time that a deep learning approach has been successfully demonstrated for the NLoS reconstruction problem using transient imaging. Additionally, our method has the benefit of being parameter-free and is highly efficient with prediction times of about 20 milliseconds.  %
\section{Deep non-line-of-sight reconstruction}

\begin{figure}[t]
	\centering
	\includegraphics[width=0.65\columnwidth]{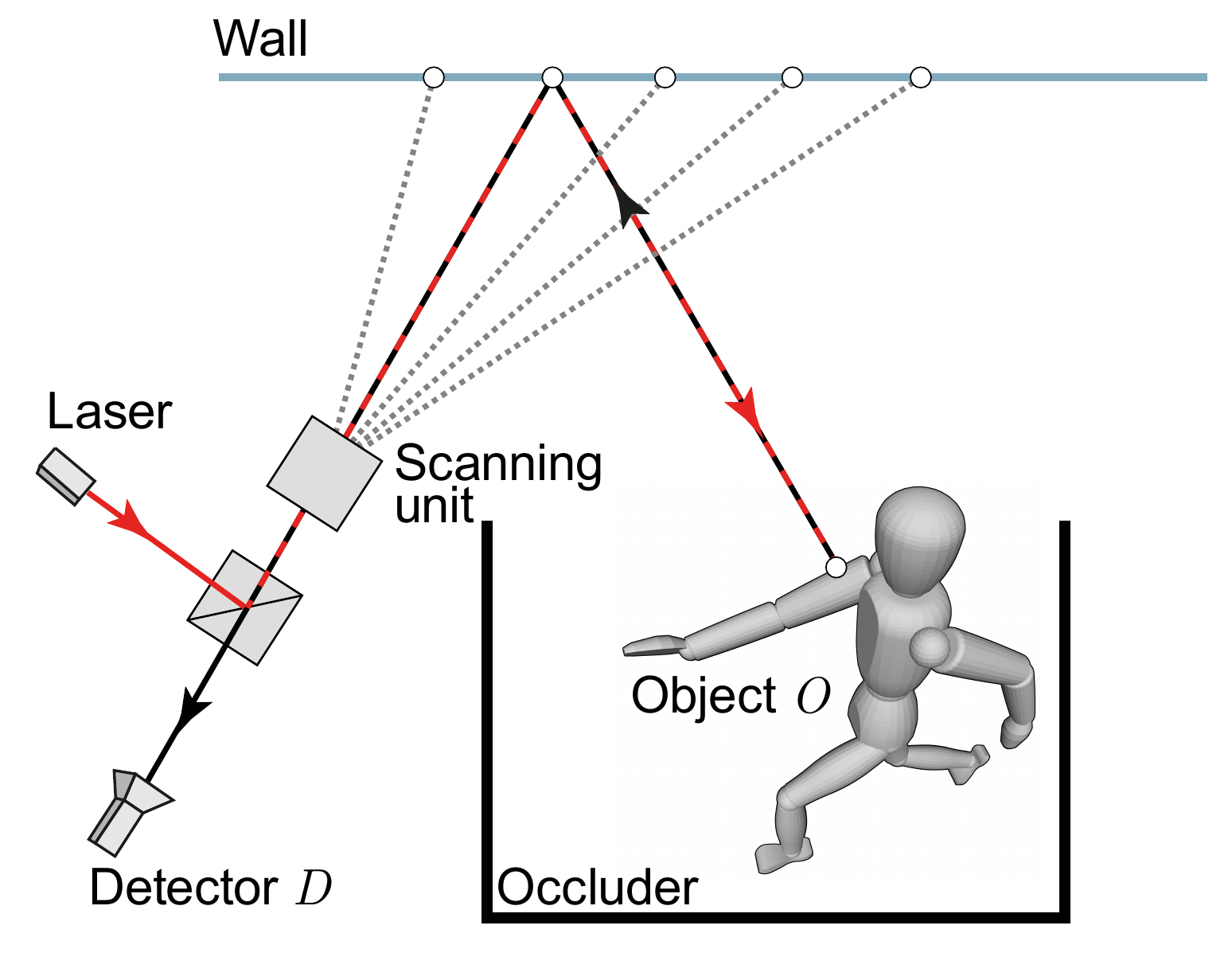}
	\caption{Non-line-of-sight capture setting. The target object $O$ is hidden behind an occluder and not directly visible by light sources and sensors. By pointing a laser source toward a wall, the occluded part of the scene can be probed with indirect light whose reflection back to the wall can then be picked up by a sensing device $D$. The arrangement depicted here is the confocal setting introduced by O'Toole et al.~\cite{otoole2018}, where a laser source and a single-pixel detector are combined through a beam splitter and scanned to an array of locations on the wall.}
	\label{fig:NLoSscene}
\end{figure}

The purpose of our work is to devise a deep learning approach to the NLoS reconstruction problem. In this section, state the problem to be solved, define the input and output of our method, and motivate a suitable deep learning architecture.

\subsection{Problem statement}
A typical sensing setup is shown in \Figure{fig:NLoSscene}. A laser source projects a very short pulse of light (often modelled as a Dirac pulse) to a location on a wall, from where it is reflected into the occluded part of the scene and, after interaction with the hidden target, picked up from the wall by a time-resolved detector. 
We formulate the operation of such a system in terms of the \emph{forward model}
$T$, which comprises
 the full image formation including light sources, propagation and detection.
The non-line-of-sight reconstruction problem relates to an \emph{inverse} mapping, where the measurement $I$ is given and the hidden target $O$ is to be estimated:
\begin{equation}\label{forward_model}
O = T^{-1}(I).
\end{equation}
This mapping $T^{-1}$ is what we aim to encode in a deep learning approach.

\subsection{Input: Transient image}
Depending on the implementation, a setup as described above could produce data in various forms ranging from a timestamped serial stream of individual photon detection events to histograms that are densely sampled in space ($x$,$y$) and time ($t$). For the purpose of this work, we assume a fully sampled space-time cube of time-resolved intensity measurements. Such a 3D volume of intensity data $I_{xyt}$, also called a \emph{transient image} \cite{smith2008transient,Velten:2012:Recovering}, will therefore serve as the input to our system.

\subsubsection{Transient Rendering}
Several techniques have been developed in recent years for rendering transient light transport~\cite{iseringhausen2018,jarabo2014framework,marco17transient,smith2008transient,tsai2019beyond,Wu2014}. Due to its accuracy and efficiency, we opt for the technique introduced by Iseringhausen et al.~\cite{iseringhausen2018}, which can be adapted to various capture geometries including the confocal setting \cite{otoole2018}. As the renderer accumulates light contributions per triangle of a meshed object, we triangulate our depth samples and feed them to the renderer. The reflectance of all surfaces is assumed to be diffuse. The resulting transient rendering is stored in the NLoS dataset with the corresponding depth map as label.

\subsubsection{Sensor model}\label{sensor_model}
The rendering algorithm discussed above does not encompass all effects that should be covered in the image formation model $T$, most prominently the sensor response of the acquisition hardware. Basic light propagation principles as well as the hardware used to implement NLoS measurements in the real world, significantly contribute to the quality of the resulting signal. 
Since most publicly available real-world datasets have been acquired by single photon avalanche detectors (SPADs), we seek to extend our forward model (\ref{forward_model}) to include the most important features of this technology. While very advanced and accurate models for SPAD sensors exist~\cite{hernandez2017computational}, they require full knowledge of the relevant parameters of the setup and are expensive to evaluate. Since we are interested in feeding networks with tens of thousands of sensor-augmented samples during training, we opt for a simplified model in favor of efficiency. Given a sample $I(x,y,t)$ generated by our renderer, we approximate the SPAD response as
\begin{equation}
I'(x,y,t) = \mathcal{P}(c \cdot I(x,y,t) + b),\label{eq:spad_model}
\end{equation}
where $c$ is a scale factor that converts unitless intensity values from the renderer to photon counts. We randomly sample $c$ from an interval that is conservatively chosen by hand and contains the correct factor with high confidence. Including this variation in the training data encourages the regressor to become invariant to global changes in intensity.
The global bias $b$ accounts for the base level of dark counts expected in transient measurements~\cite{hernandez2017computational}. To estimate it, we look at the temporal bins in real-world measurements that precede the response from the scene. Again, we randomly sample this value from an adequately chosen interval to achieve invariance with respect to this global offset.
Finally, the function $\mathcal{P}$ applies Poisson noise to reflect photon counting statistics. 

\begin{figure}[t]
	\includegraphics[width=1.02\columnwidth]{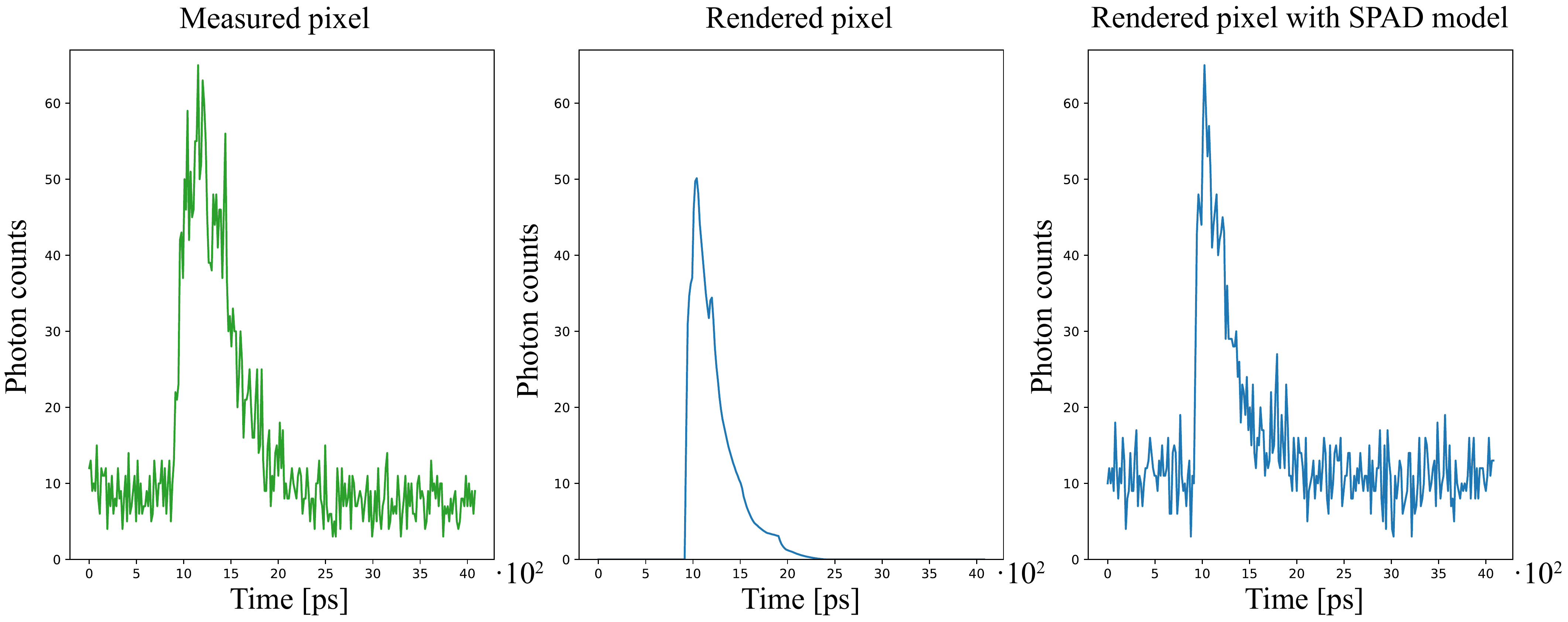}
	\caption{Comparison the time histograms of a real pixel vs.~our SPAD pixel model. {Left:} A real pixel measured with a SPAD sensor \cite{otoole2018}. {Center:} A rendered pixel computed by using the forward model \cite{iseringhausen2018}. {Right:} Rendered pixel after applying our SPAD model.}
	\label{fig:spadmodel}
\end{figure}

\Figure{fig:spadmodel} shows how our SPAD model approaches a rendered pixel to an experimental SPAD measurement. Although the model provides a coarse approximation, it captures the overall features of the real sensor while being efficient to evaluate. This enables its evaluation ``on the fly'' during training and to use its parameters for data augmentation.

\subsection{Output: Predicting depth maps}
Another central design decision relates to the representation of scene geometry. Among the candidates are volumetric (e.g., a box filled with scattering densities) and surface representations, with the latter offering a more compact encoding of opaque scenes. Since deep learning of generative models using irregularly structured geometry data is still an open problem \cite{Huang:2017:LLS:3151031.3137609,Achlioptas2018,Mescheder2019} and volumetric representations tend to be highly memory-intensive during the training phase, we decide to represent the scene as a depth map (a 2D map of range values). Choosing this model as output of our system comes with the sacrifice of not being able to reflect certain types of self-occlusion~\cite{Heide:2019:OcclusionNLoS}. Yet, the representation proves reasonably versatile for a wide range of test scenes as shown, for instance, in a recent optimization framework~\cite{tsai2019beyond}.

\begin{figure}%
\includegraphics[width=\columnwidth]{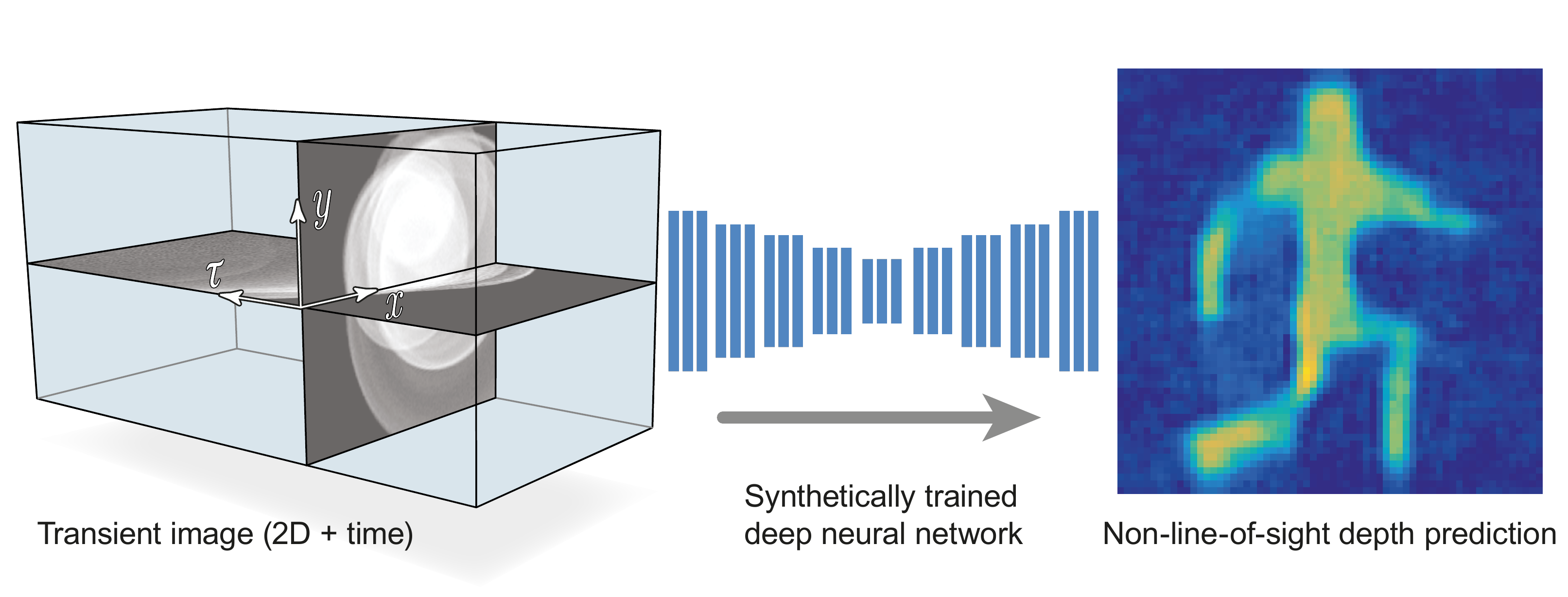}%
\caption{Schematic overview of the reconstruction pipeline. Input is provided in the form of a transient data volume. A convolutional neural network maps this data to a 2D depth map.}%
\label{fig:pipeline-overview}%
\end{figure}

\subsection{Deep learning architecture}
Given these design choices regarding the input and output interfaces, the resulting high-level data flow is illustrated in \Figure{fig:pipeline-overview}. Our deep learning approach employs a synthetically trained neural network that retrieves a hidden target from indirect measurements of the scene. Modelling such inverse problems using end-to-end trained convolutional regressors has proven successful for a wide variety of computer vision tasks on dense data. In classical vision problems like image segmentation, a 1-to-1 mapping of an input image to per-pixel object class weights is desired. One possible architecture is an autoencoder, which encodes a representation of the input data by performing dimensionality reduction, and then unpacks it back to the original format~\cite{girdhar2016learning}. An approach to learning input-to-output mappings consists in tailoring architectural features of such network and train in a supervised way~\cite{Wei2016}. Such a neural regressor, as we will use in our approach, is naturally limited to the entropy of the input data. If further details should be hallucinated, generative statistical modeling has to be used, for example through variational autoencoders or generative adversarial networks~\cite{goodfellow2014generative,radford2015unsupervised}. We do not follow this path, as we seek objective rather than visually pleasing reconstructions.

Our setting is characterized by a dimensionality mismatch: from a 3-dimensional input (two spatial dimensions plus time of flight), we want to infer a 2-dimensional output (a depth map). While there can be some correlation between spatial dimensions in the input and the output \cite{otoole2018}, this does not necessarily hold in all sensing geometries. Similar problems have for example been studied in deep network-based reconstruction of 3D images from 2D views \cite{Dosovitskiy2017,girdhar2016learning,Lin2018learning,sitzmann2019deepvoxels,Wu2015CVPR}. Where most prior work tries to infer 3D output from 2D input, our challenge is to go from the 3D transient image to 2D depth map output and hence more closely related to work on SPAD data denoising \cite{Lindell:2018:3D}. While sophisticated translation mechanisms are possible, such as sparse point clouds and differentiable rendering~\cite{Lin2018learning}, a simple solution is to use densely connected networks as a regressor mechanism~\cite{girdhar2016learning}. Our architecture combines convolutional layers with fully connected columns in time with a final enhanced decoder for optimal reconstruction quality at reasonable cost. To our knowledge, no prior work has proposed an end-to-end deep NLoS reconstruction pipeline.

\begin{figure*}[t]
\centering
\includegraphics[width=0.9\textwidth]{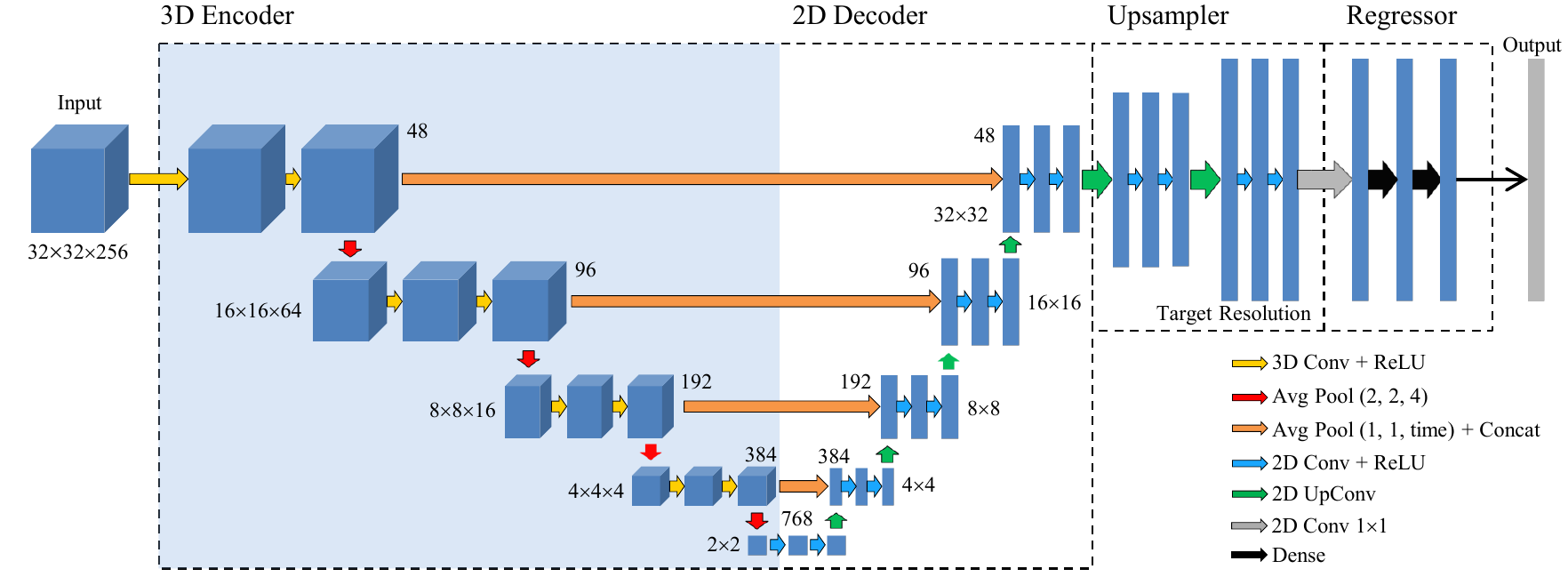}
\caption{Our functional model for the NLoS reconstruction task. Time-of-flight volumes are analyzed by a 3D$\rightarrow$2D encoder-decoder in the first stage. Resulting features are upsampled to a specified target resolution and further decoded into the object's representation, in this case a depth image of the hidden region.}
	\label{fig:neuralnetwork}
\end{figure*}

A detailed overview of our deep neural network is shown in \Figure{fig:neuralnetwork}. The objective of our design is to compress the 3D input in the first two stages, which are to be decoded as 2D depth maps by a fully-connected regressor in the third stage. The latter is important; we have experimentally observed that leaving out any densely-connected mechanism increases the reconstruction error by about 10\%. We motivate three main components within our network as follows:

\noindent\textbf{3D/2D convolutional autoencoder.}\quad At the entry level of the network, we construct an encoder-decoder network with skip connections \cite{ronneberger2015unet,cicek20163d}. On the encoder side, the network contracts spatial and temporal dimensions of the transient image, while increasing the number of convolutional channels at each layer. Feature maps are extracted using 3D convolutions of size 3$\times$3$\times$9, which we gradually diminish by 2 along the temporal size after each pooling. We chose average downsampling over max-pooling for compressing our inputs as the former showed better validation performance when training on small batches. On the other side, the decoder network implements 2D up-convolutions of constant size 2$\times$2 to upsample 2-dimensional maps from the latent vectors contracted during encoding. Skip connections from the encoder to the decoder to enhance gradient propagation and scale separation.

\noindent\textbf{Upsampler network.}\quad  Once the U-Net has again reached the spatial resolution of the input, we stack an upsampling network that expands incoming maps into higher spatial resolutions. Among the choices available, we opt for the same upsampling mechanism as the decoder, namely an up-convolution followed by two 2D convolution + ReLU layers.

\noindent\textbf{Regressor network.}\quad  When the upsampler has reached a specified target resolution, we stack a regressor network that computes depth images from the feature maps. It consists of a 1$\times$1 convolution that contracts incoming channels, and two fully-connected layers.

\subsection{Training strategy}\label{neural_model}
We implement our system in Keras with TensorFlow as the backend. To train our network, we backpropagate the mean squared difference of depth maps. We experimented with $L_1$ losses, but experiments have shown instabilities during training and / or convergence. Weights are initialized as in \cite{He_initialization} and optimized using Adam with default settings and a learning rate of 0.0001. The training was performed on an Nvidia GeForce RTX 2080 Ti over a runtime of 3--5 days, and the models were regularized by means of early stopping. %
The long training times and the need for regularization are mostly due to the fully-connected regressor at the end of our pipeline, which contains most of the parameters. We opted for early stopping at the minimal validation error rather than drop-out due to better performance in our experiments.
\begin{figure}[t]
\centering
\rotatebox{90}{\small~~~FlatNet\textcolor{white}p} 
\includegraphics[trim=11mm 8mm 5mm 3mm,clip,width = 0.18\columnwidth]{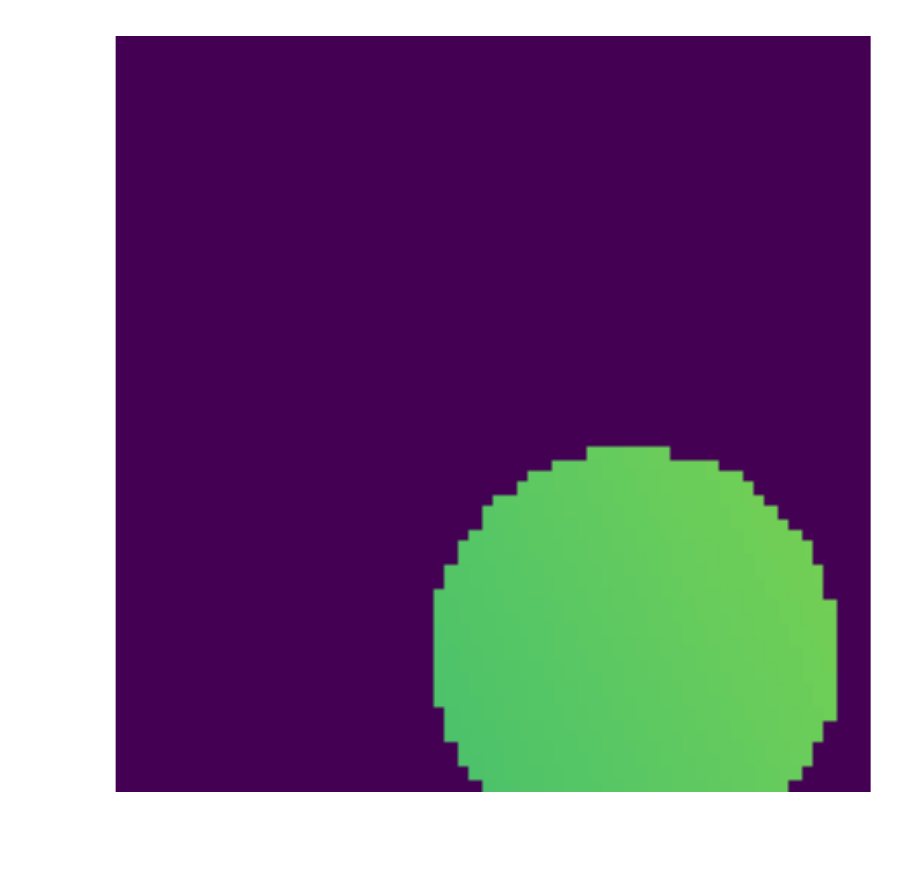}%
\hfill
\includegraphics[trim=11mm 8mm 5mm 3mm,clip,width = 0.18\columnwidth]{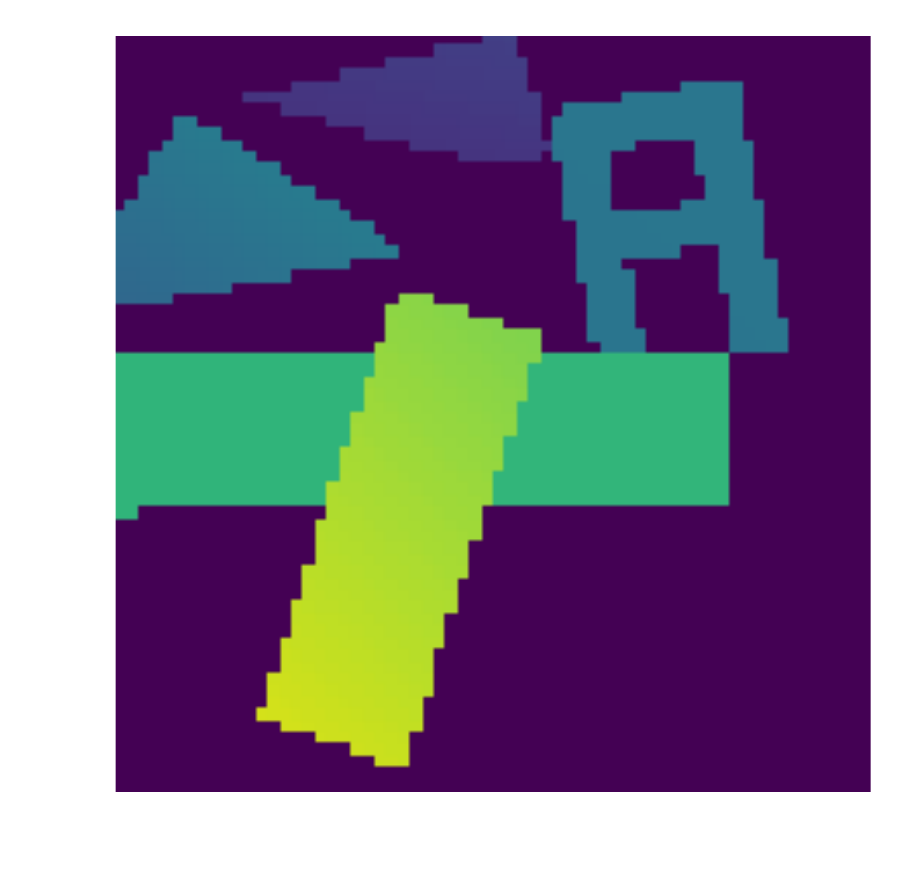}%
\hfill
\includegraphics[trim=11mm 8mm 5mm 3mm,clip,width = 0.18\columnwidth]{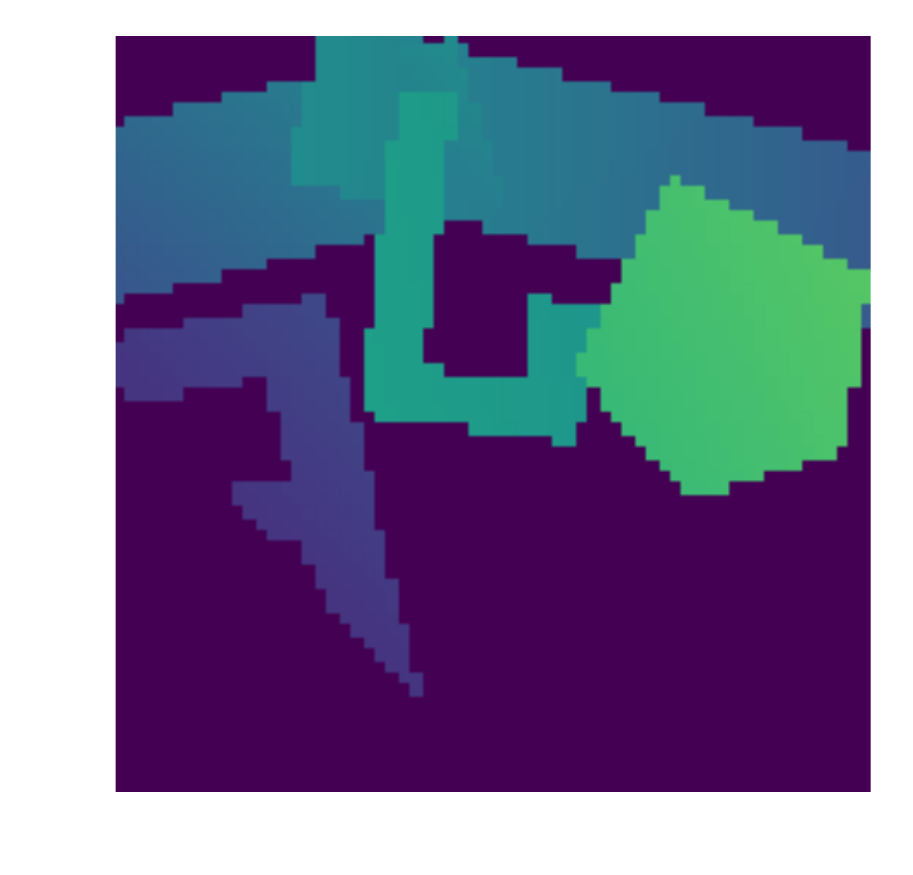}%
\hfill
\includegraphics[trim=11mm 8mm 5mm 3mm,clip,width = 0.18\columnwidth]{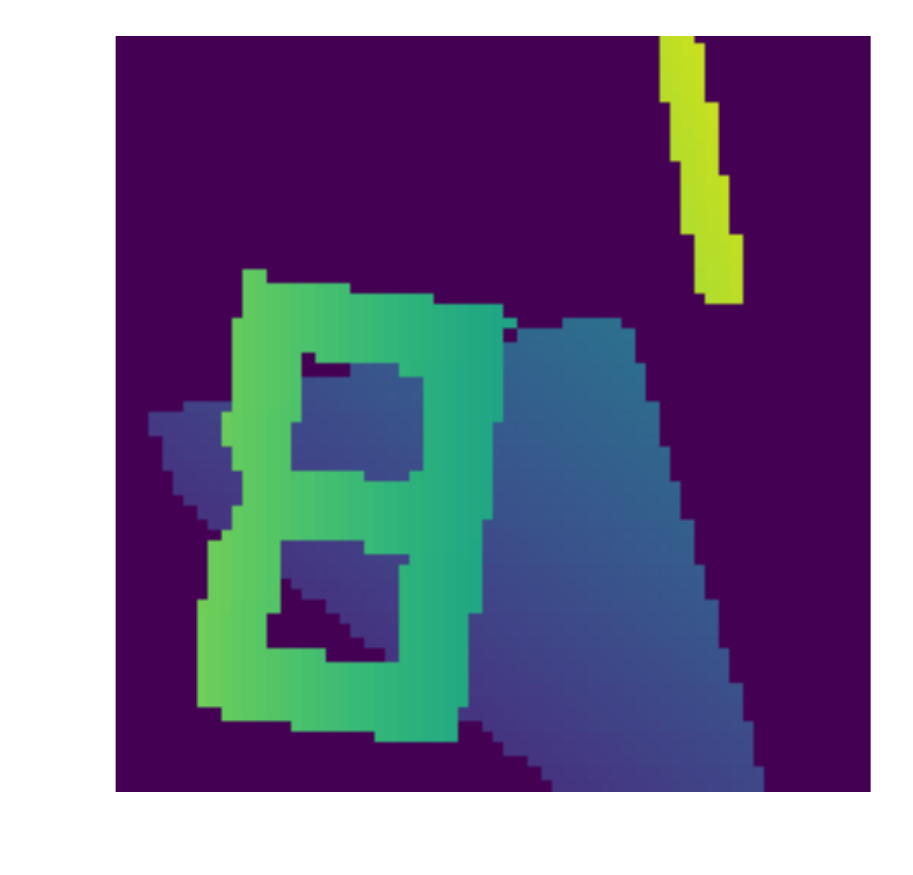}%
\hfill
\includegraphics[trim=11mm 8mm 5mm 3mm,clip,width = 0.18\columnwidth]{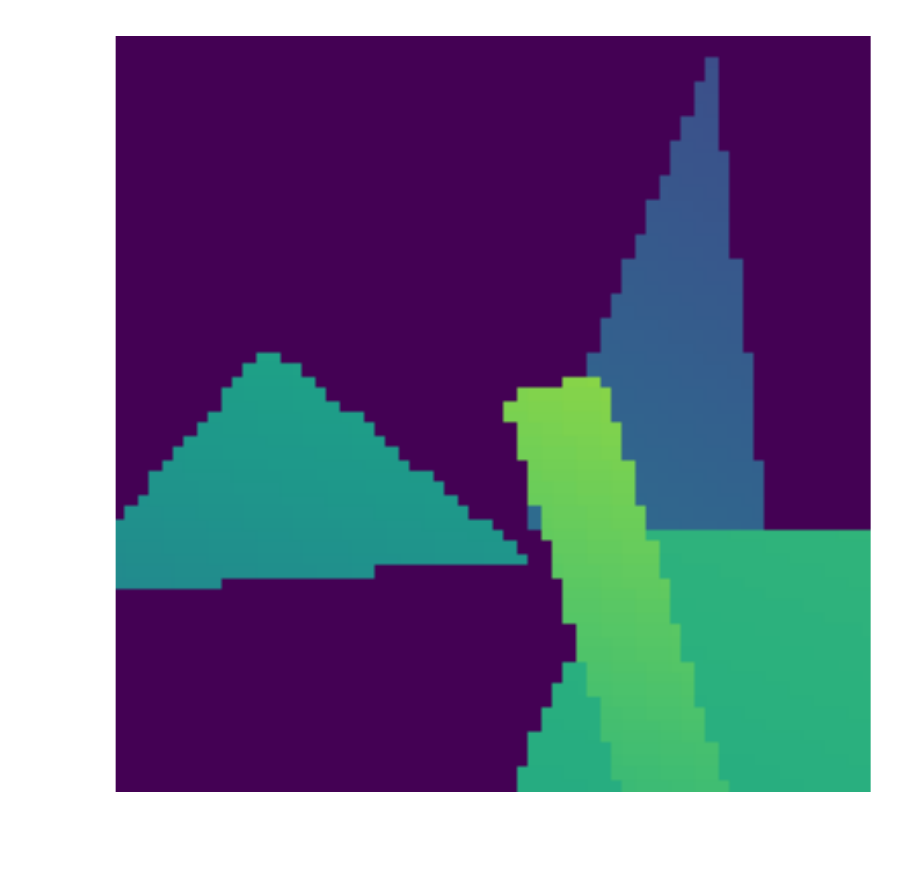}\\[0.5mm]
\rotatebox{90}{\small~~ShapeNet\textcolor{white}p} 
\includegraphics[trim=11mm 8mm 5mm 3mm,clip,width = 0.18\columnwidth]{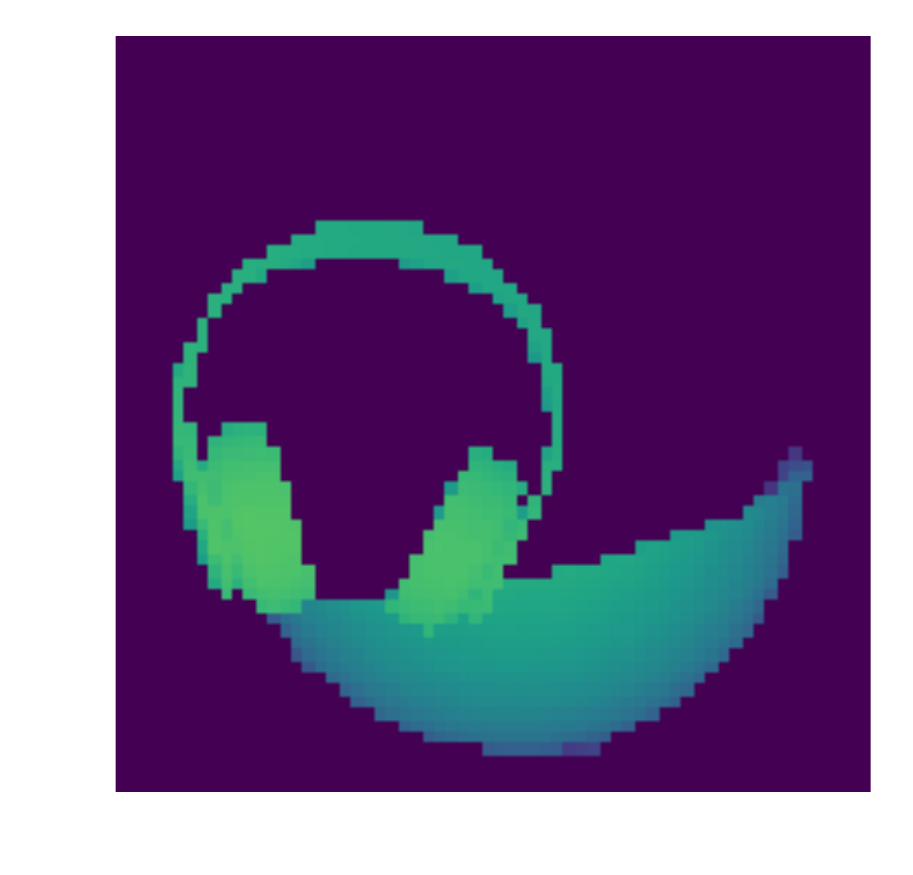}%
\hfill
\includegraphics[trim=11mm 8mm 5mm 3mm,clip,width = 0.18\columnwidth]{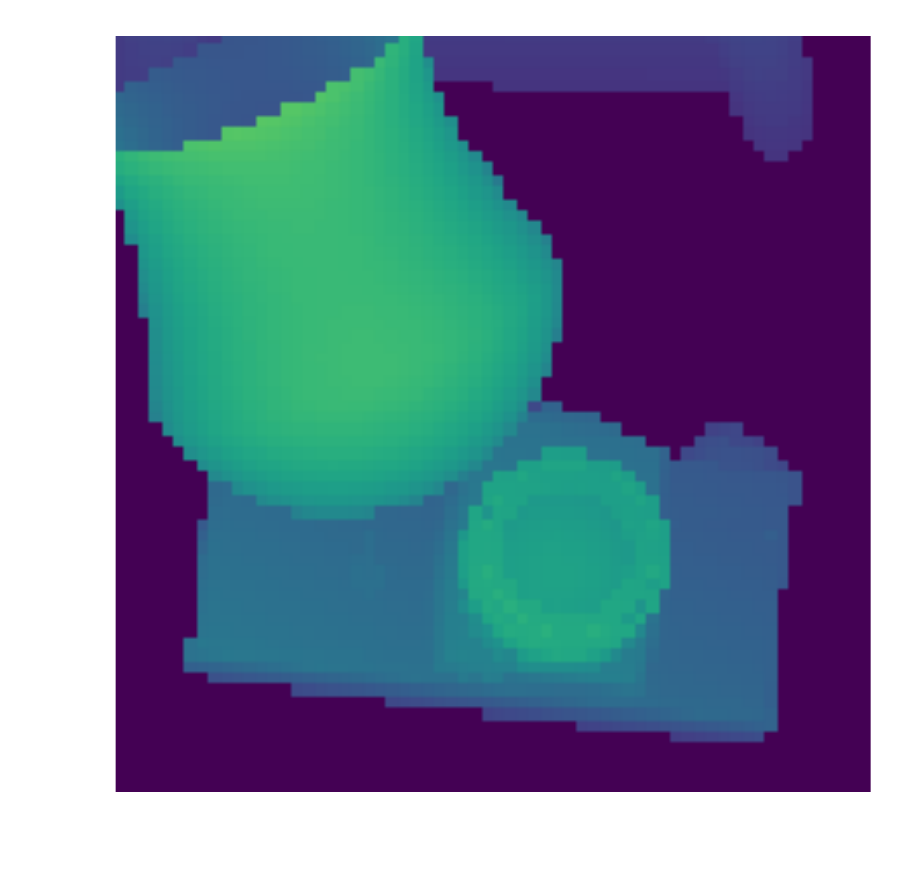}%
\hfill
\includegraphics[trim=11mm 8mm 5mm 3mm,clip,width = 0.18\columnwidth]{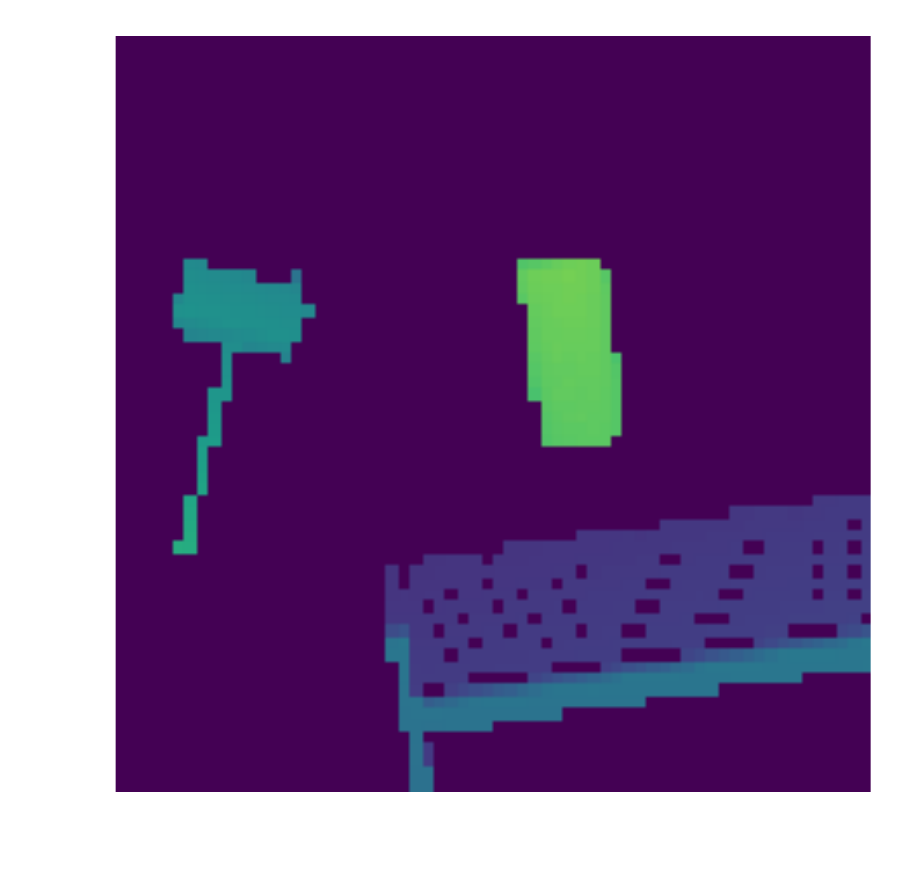}%
\hfill
\includegraphics[trim=11mm 8mm 5mm 3mm,clip,width = 0.18\columnwidth]{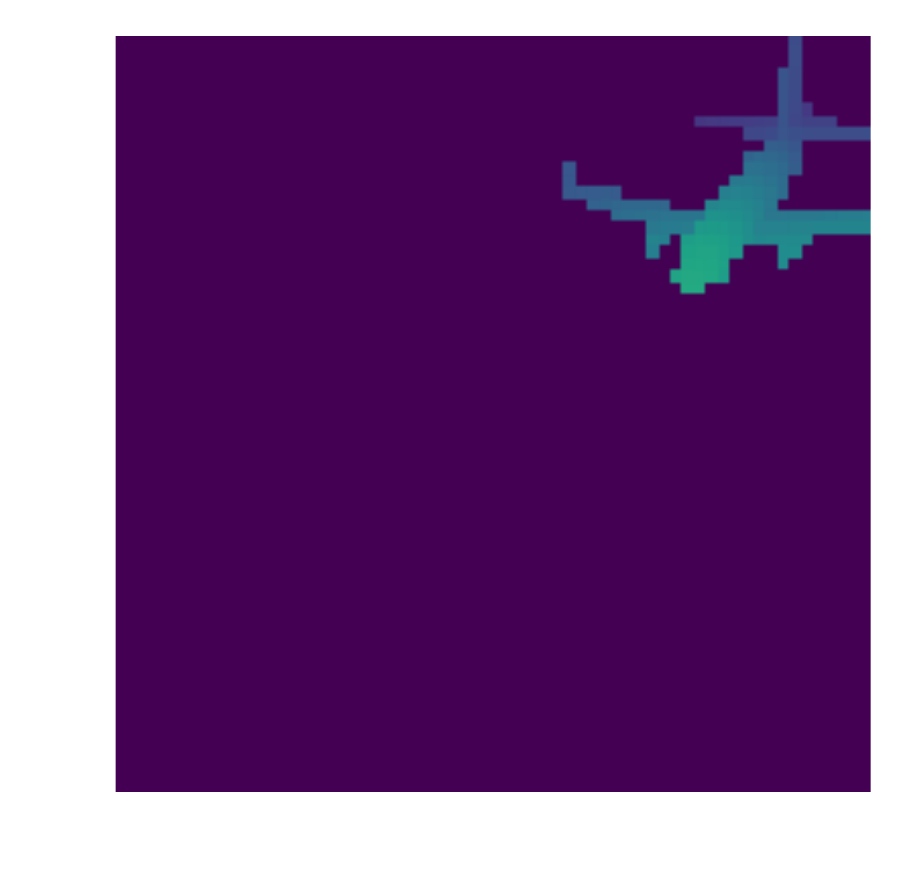}%
\hfill
\includegraphics[trim=11mm 8mm 5mm 3mm,clip,width = 0.18\columnwidth]{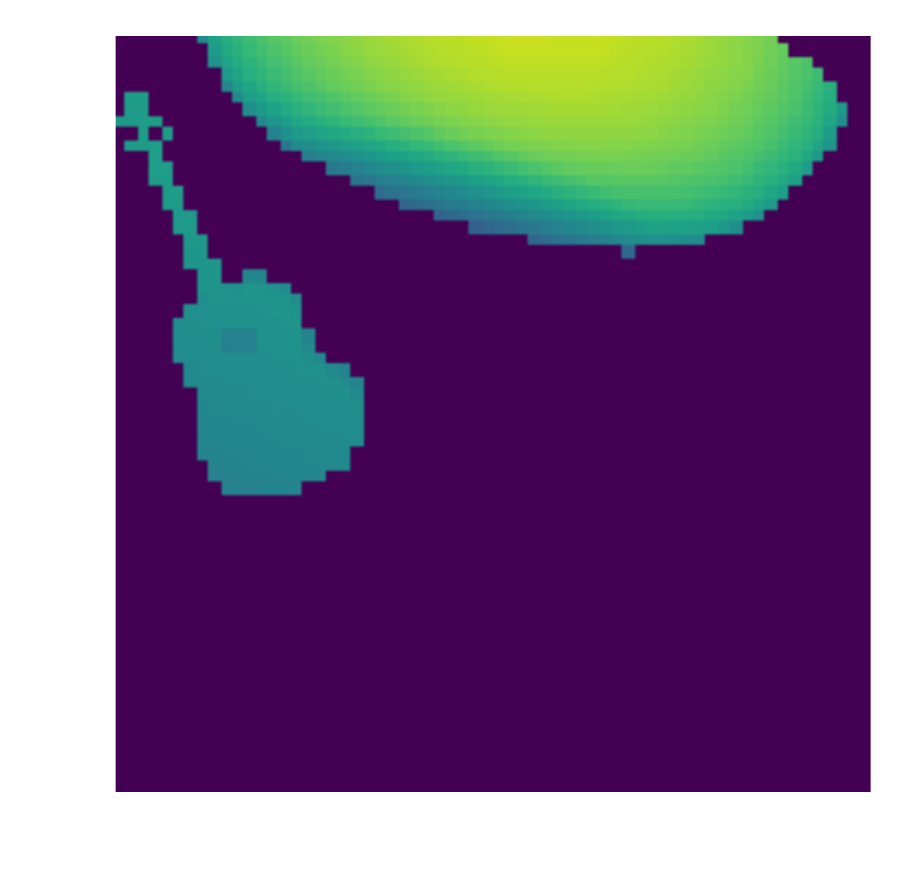}\\[0.5mm]
\rotatebox{90}{\small~~Redwood\textcolor{white}p} 		
\includegraphics[trim=11mm 8mm 5mm 3mm,clip,width = 0.18\columnwidth]{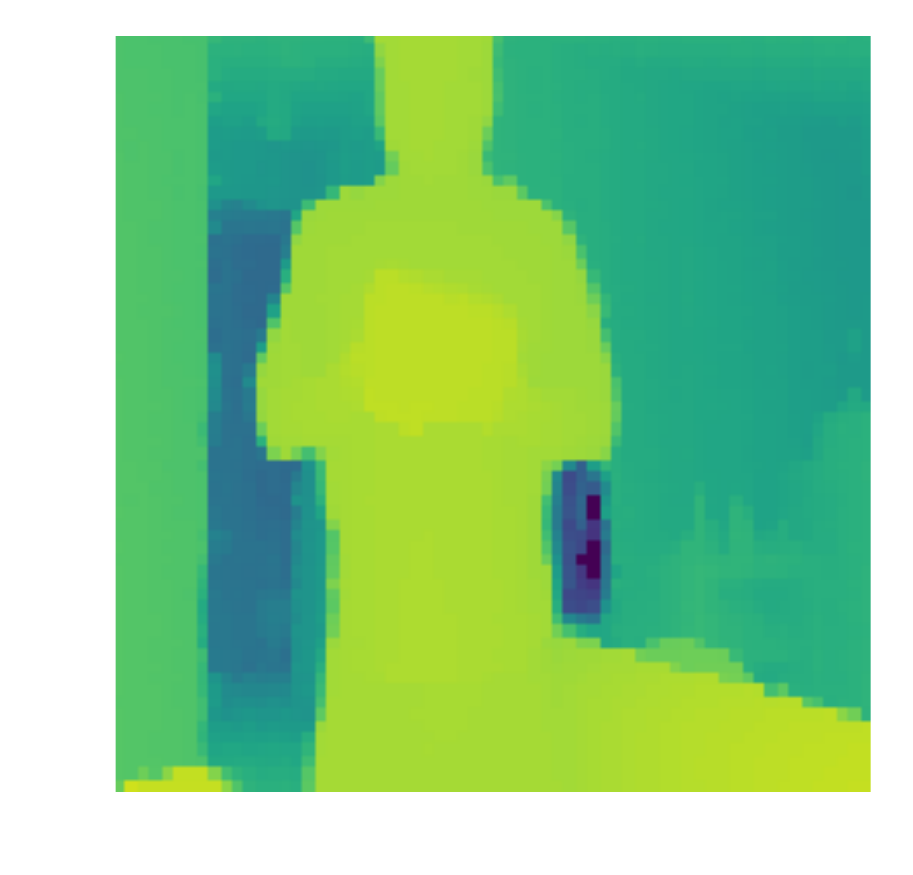}%
\hfill
\includegraphics[trim=11mm 8mm 5mm 3mm,clip,width = 0.18\columnwidth]{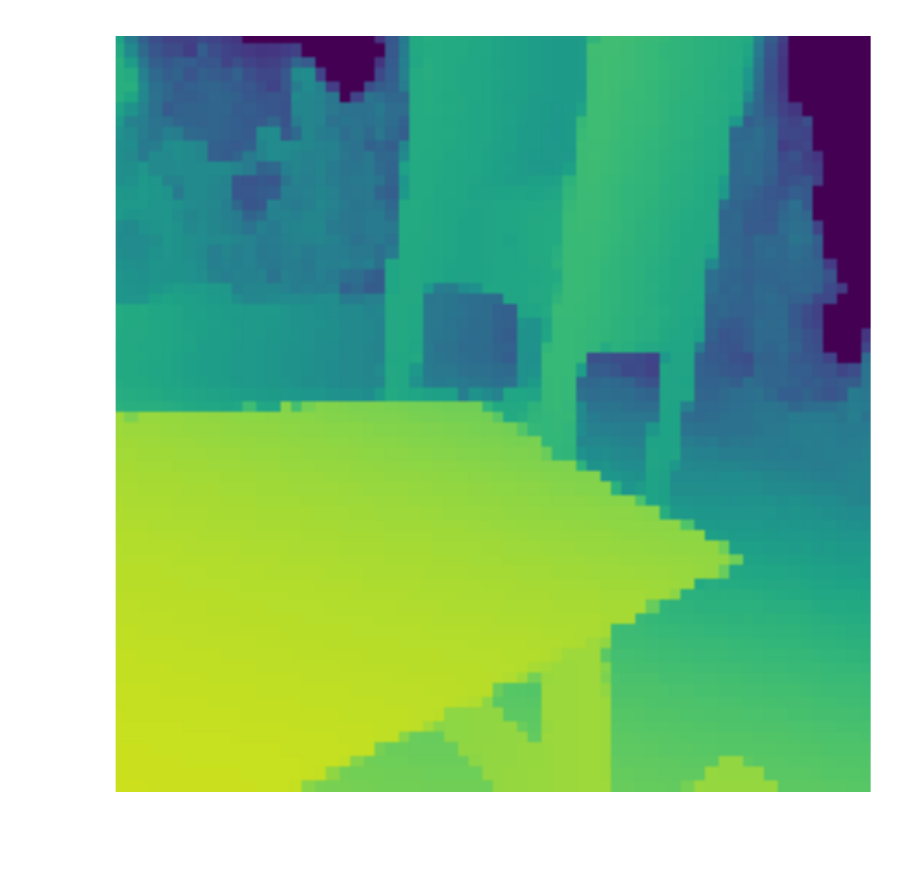}%
\hfill
\includegraphics[trim=11mm 8mm 5mm 3mm,clip,width = 0.18\columnwidth]{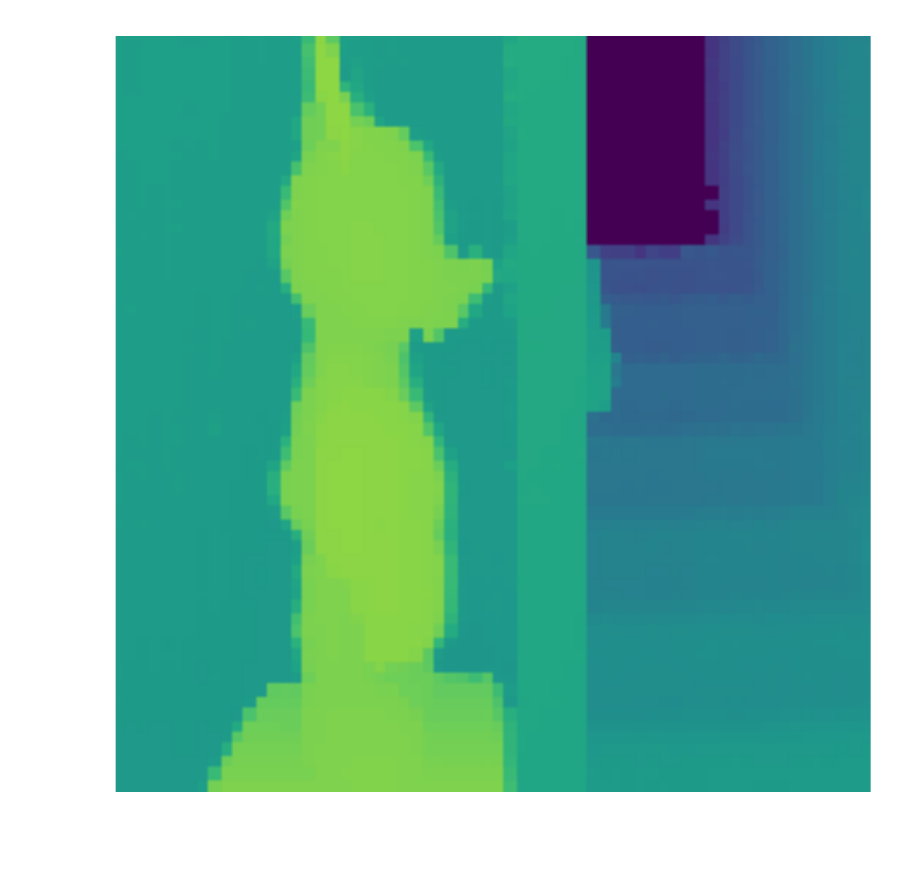}%
\hfill
\includegraphics[trim=11mm 8mm 5mm 3mm,clip,width = 0.18\columnwidth]{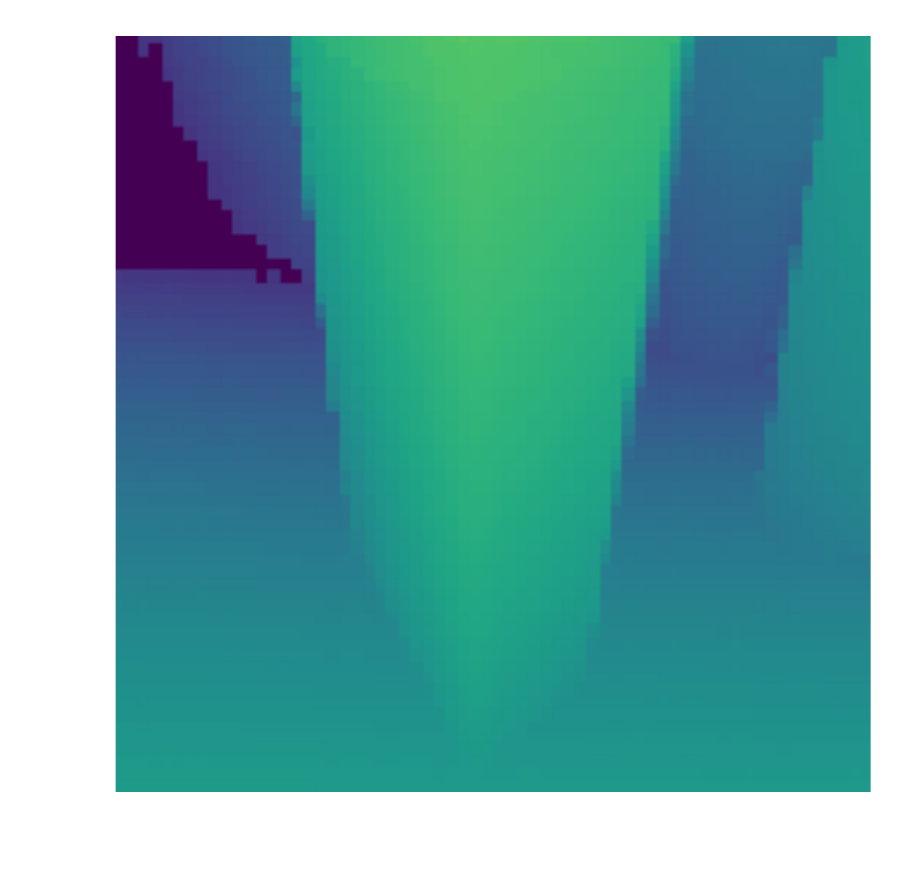}%
\hfill
\includegraphics[trim=11mm 8mm 5mm 3mm,clip,width = 0.18\columnwidth]{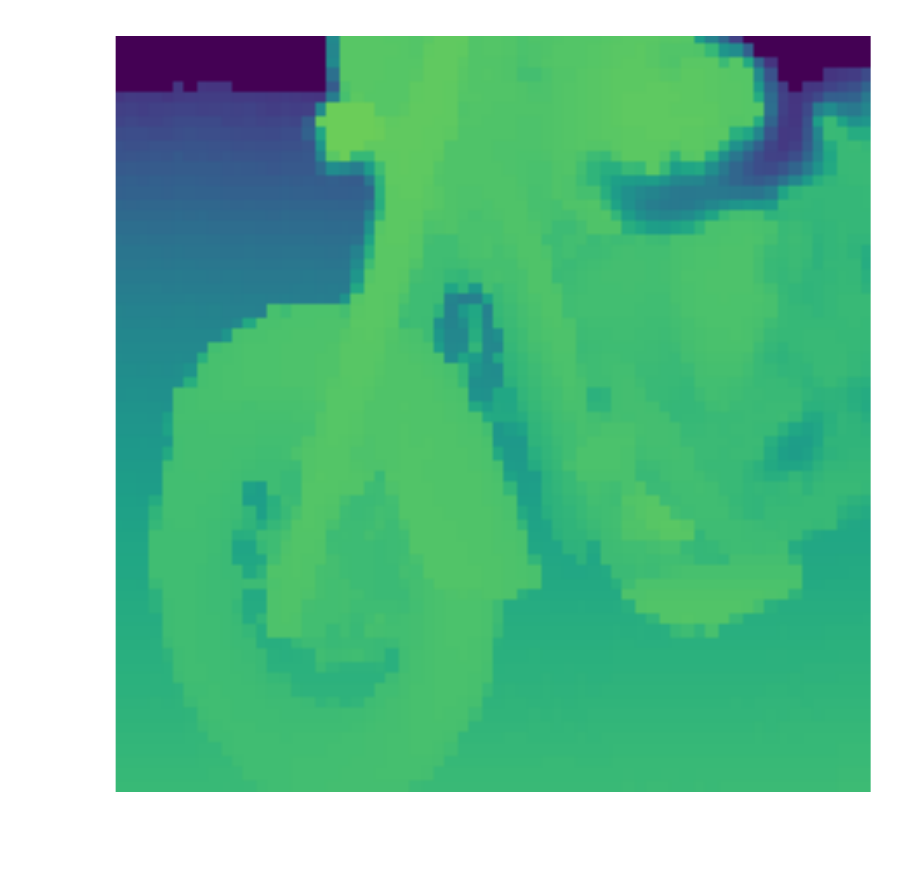}\\[-0.5mm]
	\caption{Examples of depth maps used for training. The three datasets exhibit significantly different foreground / background statistics.}
	\label{fig:datasets}
\end{figure}

\section{Results and evaluation}
In this section, we describe how we generated datasets for our method and evaluate our system on synthetic and real test cases.

\subsection{Training models on synthetic data}
The first step of our deep learning pipeline consists in acquiring depth map instances to train our network. In general, depth map collections coming from different sources exhibit distinct foreground vs. background statistics, which in turn endorses neural models with different predictive capabilities. In this section, we seek to explore such distinctions by training models over depth map datasets that have been collected from either one mechanism or source at a time. By keeping separate datasets of different statistics, we hope to provide useful insights for particular application domains.
To build depth map datasets, we consider the cases of collecting from either synthetic sources or real-world measurements. In the synthetic case, given a database of 3D models, we generate a depth sample by placing one or more models inside a bounding volume. After positioning each model in the scene with a random affine transformation, the resulting depth map is generated by applying an orthographic projection and then storing the z-buffer values. We build two datasets with this strategy, FlatNet and ShapeNet, which differ only by the model database used. The first uses strictly flat shapes as source models (circles, triangles, letters, etc), while the second samples shapes from the ShapeNetCore database \cite{shapenet:paper}.

For the case of real-world depth map datasets, we extract crops of randomly sampled depth frames from the publicly available RGB-D datasets, in our case the Redwood database \cite{redwood2016:paper}. In this case, we found some pre-processing choices to have an impact on the final dataset statistics (foreground/background), and therefore considered the following heuristics: 1) the dataset should be artifact-free, 2) rich class variability. To proceed, we uniformly sampled categories in order to reduce class imbalance. When addressing noise, we remove missing pixels and border artifacts. We found that, in general, this a difficult task to achieve across the entire dataset as captures not only possess different resolutions, but also different noise according to several scene factors (indoors, outdoors, motion, etc). We inpainted regions of invalid depth (depth $\le0$) using diffusion. Then, we extracted square crops centered at the image center and downsampled them to the network's resolution.

Examples from the three datasets can be seen in \Figure{fig:datasets}. Each depth map is transformed to the bounding volume $[-1,1]^3$, where $z=-1$ represents the background plane and $-1<z\leq1$ foreground geometry. For each dataset, we generate 40,000 depth maps of 64$\times$64 pixels resolution and render the corresponding transient images. The resulting volumes are downsampled to 32$\times$32$\times$256 pixels, with an effective time resolution of 16\,ps. Datasets are partitioned into 36,000 training pairs and 4,000 test pairs.

\subsection{Evaluation on synthetic transient images}
\begin{figure}
	\includegraphics[width=\columnwidth]{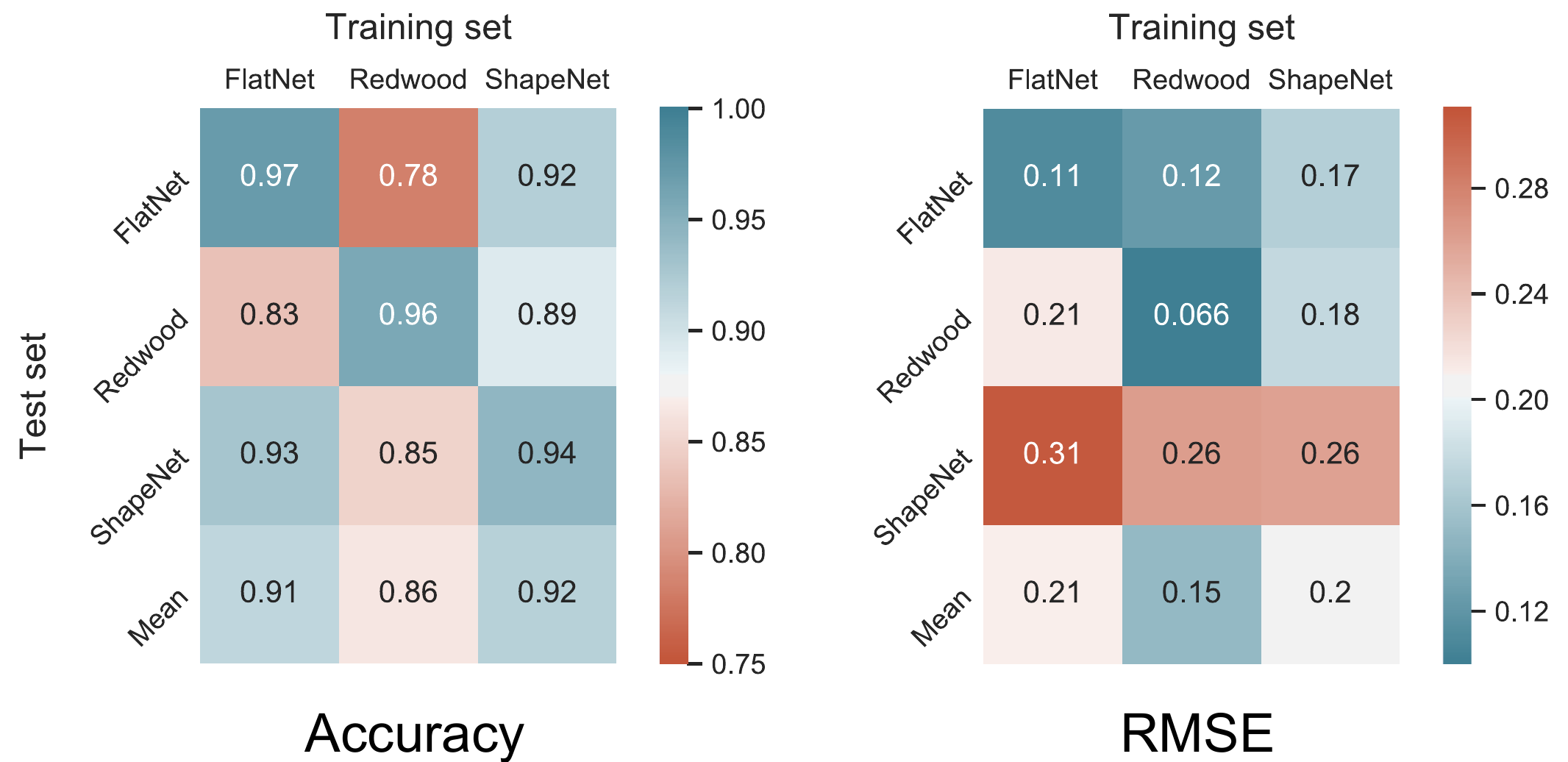}
	\caption{Average accuracy and RMS depth error across datasets. Columns are trained with the same dataset, while rows are tested using the same dataset. Models trained with synthetically generated depth maps (FlatNet and ShapeNet) exhibit better mean accuracy (correct geometry) than the model trained on RGB-D data (Redwood). On the other hand, the Redwood model exhibits better at RMSE scores of the true positive pixels, where FlatNet and ShapeNet perform comparably.}\label{fig:accuracy}\label{fig:rmse}
\end{figure}

When evaluating performance we consider two important aspects of the reconstruction task. On one hand, we note that an evaluation metric must account for the quality of foreground / background classification. We use the classification accuracy
\begin{equation}
\textrm{Accuracy}=\frac{\textrm{TP}+\textrm{TN}}{\textrm{TP}+\textrm{TN}+\textrm{FP}+\textrm{FN}},
\label{eq:accuracy}
\end{equation}
combining rates for missing geometry (false negatives / FN), excess (false positives / FP), correct foreground (true positives / TP) and background (true negatives / TN). In addition, we employ a root-mean-square error (RMSE) measure to evaluate the depth error for true positives. \Figure{fig:accuracy} shows the cross-validated errors on the test partitions of each dataset. From this, we observe that the synthetically-trained models, FlatNet and ShapeNet, perform statistically similar in both, accuracy and RMSE terms. On the other hand, the real-data model, Redwood, outperforms the latter two in the RMSE sense but exhibits poorer accuracy than these, manifested as overestimated depth pixels. Unlike FlatNet and ShapeNet (with 71\% and 78\% background pixels, respectively), Redwood only contains 7\% background pixels across the entire database. Such statistics could likely explain the observed similarities between the synthetically trained models and the discrepancies between these and the real-data model.

\begin{figure}[t!]
	\centering
	\rotatebox{90}{\textcolor{white}{Gp}} %
	\begin{minipage}[c]{.25\columnwidth}
	\centering \small SynthBunny
	\end{minipage}
	\begin{minipage}[c]{.25\columnwidth}
	\centering \small SynthDiffuseS
	\end{minipage}
	\begin{minipage}[c]{.25\columnwidth}
	\centering \small \!SynthMannequin
	\end{minipage}\!\!\!\\
	
\rotatebox{90}{~~\small Ground truth\textcolor{white}p} 
\includegraphics[width = 0.24\columnwidth]{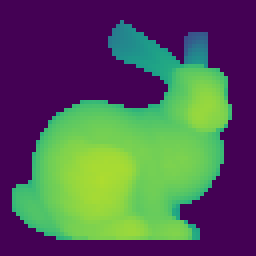} %
\includegraphics[width = 0.24\columnwidth]{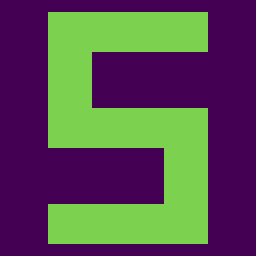} %
\includegraphics[width = 0.24\columnwidth]{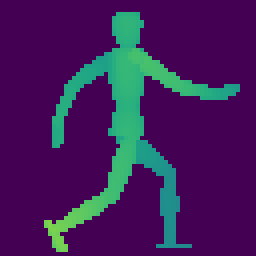}\!\!\!\\[0.2mm]%

\rotatebox{90}{~~~~~\,\small FlatNet\textcolor{white}p} 
\includegraphics[width = 0.24\columnwidth]{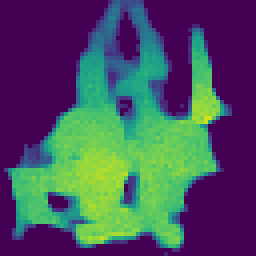} %
\includegraphics[width = 0.24\columnwidth]{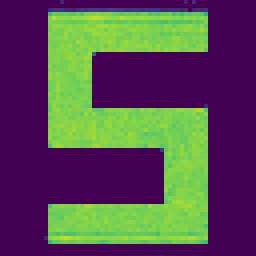} %
\includegraphics[width = 0.24\columnwidth]{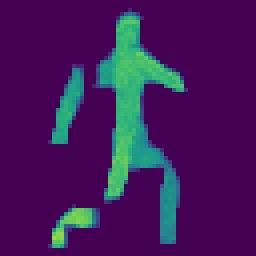}\!\!\!\\[0.2mm]
	
\rotatebox{90}{~~~~\small ShapeNet\textcolor{white}p}
\includegraphics[width = 0.24\columnwidth]{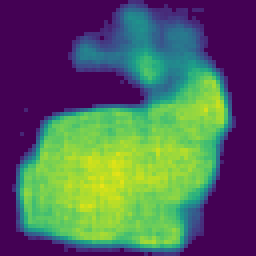} %
\includegraphics[width = 0.24\columnwidth]{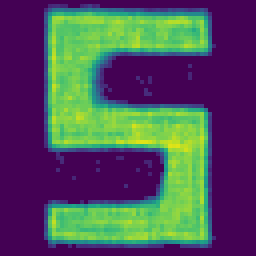} %
\includegraphics[width = 0.24\columnwidth]{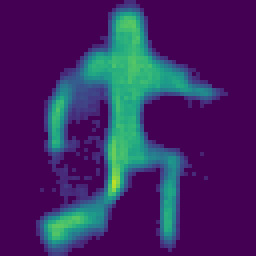}\!\!\!\\[0.2mm]
	
\rotatebox{90}{~~~~\,\small Redwood\textcolor{white}p} 
\includegraphics[width = 0.24\columnwidth]{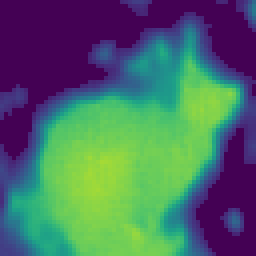} %
\includegraphics[width = 0.24\columnwidth]{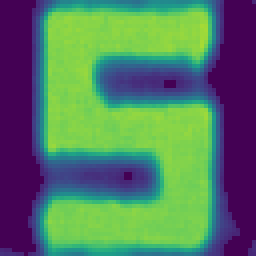} %
\includegraphics[width = 0.24\columnwidth]{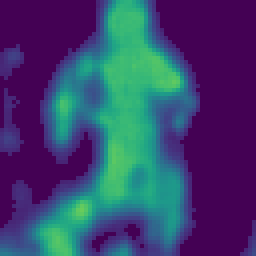}\!\!\!\\[0.2mm]

\rotatebox{90}{~~~~~~~\,\small LCT\textcolor{white}p} 
\includegraphics[width = 0.24\columnwidth]{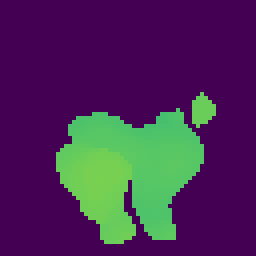} %
\includegraphics[width = 0.24\columnwidth]{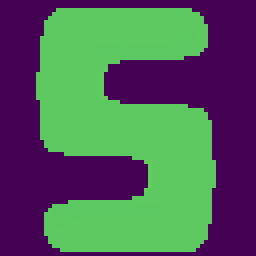} %
\includegraphics[width = 0.24\columnwidth]{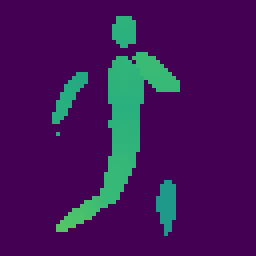}\!\!\!\\[0.2mm]
	\caption{Qualitative visualization of three synthetic depth maps as reconstructed by our architecture using different training datasets, and by LCT \cite{otoole2018}. In all cases, inputs are contaminated with Poisson noise. Reconstructions performed by our three models capture general shape features of the test cases, while exhibiting different predictive behavior. The retrieved shapes are comparable to state-of-the-art methods (LCT) in most cases.%
	}
	\label{fig:synth_eval_preds}
\end{figure}

In addition to these large-scale quantitative experiments, we manually prepare three synthetic test cases: \emph{SynthBunny}, \emph{SynthMannequin} and \emph{SynthDiffuseS}. \Figure{fig:synth_eval_preds} shows predictions of our trained models in line with the light-cone transform method (LCT) from \cite{otoole2018}. Input images in these cases have been augmented by a plausible amount of sensor noise and bias. Note that in most cases the trained models have retrieved the targets within recognizability, while being qualitatively comparable with LCT. As expected, the distinct generalization behavior of each training dataset is also visible in the predicted depth maps.

\begin{table}[t]
\caption{Depth errors and accuracy values corresponding to \Figure{fig:synth_eval_errors}.}
\label{tbl:synth_error}
\begin{center}
\begingroup
\fontsize{8pt}{10pt}
\selectfont
\begin{tabular}{c|cc|cc|cc}
& \multicolumn{2}{c|}{SynthBunny} & \multicolumn{2}{c|}{SynthDiffuseS} & \multicolumn{2}{c}{SynthMannequin}\\
Method & RMSE & Acc. & RMSE & Acc. & RMSE & Acc.\\
\hline
FlatNet & 0.32 & 0.82 & 0.06 & 0.99 & 0.24 & 0.94\\
ShapeNet & 0.21 & 0.87 & 0.20 & 0.93 & 0.32 & 0.88\\
Redwood & 0.23 & 0.79 & 0.04 & 0.67 & 0.19 & 0.69\\
LCT & 0.03 & 0.76 & 0.003 & 0.89 & 0.11 & 0.93\\
\hline
\end{tabular}
\endgroup
\end{center}
\end{table}

\begin{figure}[ht]
\parbox{\columnwidth}{
\newlength{\imagewidth}
\setlength{\imagewidth}{0.28\columnwidth}
\setlength{\tabcolsep}{.5mm}
\begin{center}
\begin{tabular}{cccc}
\rotatebox[origin=c]{90}{FlatNet} &
\raisebox{-0.5\height}{\includegraphics[width=\imagewidth]{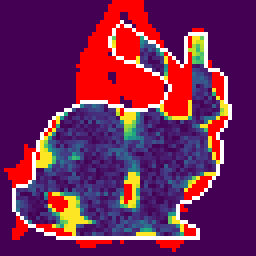}} &
\raisebox{-0.5\height}{\includegraphics[width=\imagewidth]{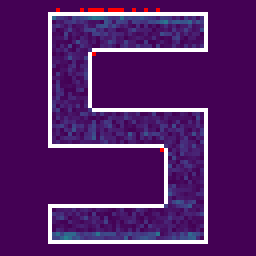}} &
\raisebox{-0.5\height}{\includegraphics[width=\imagewidth]{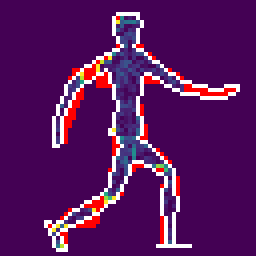}} \\[11.25mm]
\rotatebox[origin=c]{90}{ShapeNet} &
\raisebox{-0.5\height}{\includegraphics[width=\imagewidth]{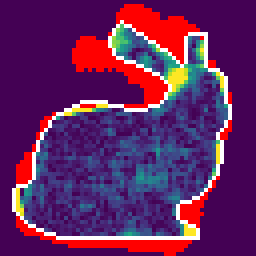}} &
\raisebox{-0.5\height}{\includegraphics[width=\imagewidth]{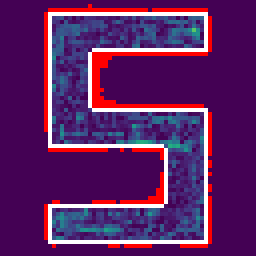}} &
\raisebox{-0.5\height}{\includegraphics[width=\imagewidth]{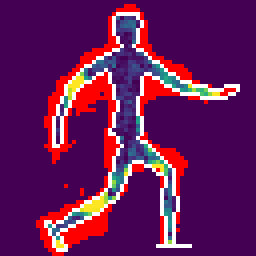}} \\[11.25mm]
\rotatebox[origin=c]{90}{Redwood} &
\raisebox{-0.5\height}{\includegraphics[width=\imagewidth]{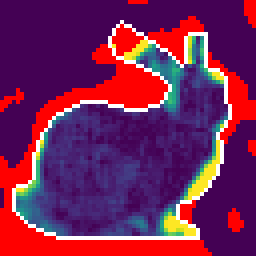}} &
\raisebox{-0.5\height}{\includegraphics[width=\imagewidth]{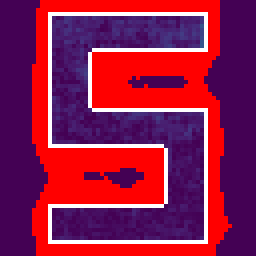}} &
\raisebox{-0.5\height}{\includegraphics[width=\imagewidth]{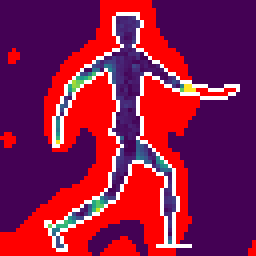}} \\[11.25mm]
\rotatebox[origin=c]{90}{LCT} &
\raisebox{-0.5\height}{\includegraphics[width=\imagewidth]{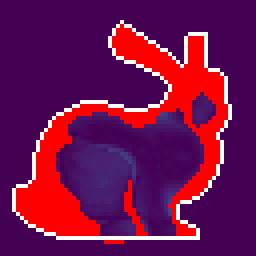}} &
\raisebox{-0.5\height}{\includegraphics[width=\imagewidth]{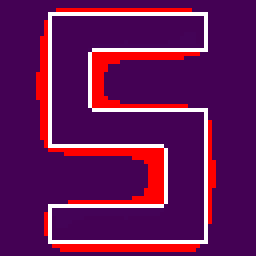}} &
\raisebox{-0.5\height}{\includegraphics[width=\imagewidth]{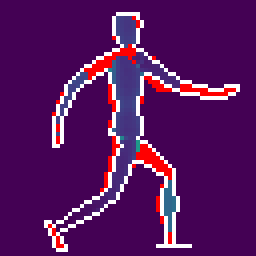}} \\[11mm]
& \multicolumn{3}{c}{\includegraphics[width=.8\columnwidth]{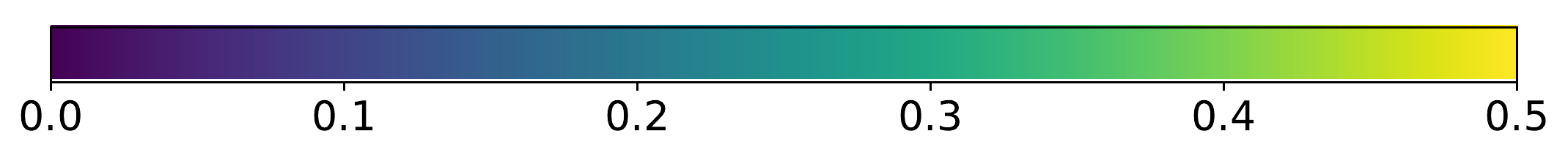}}
\end{tabular}
\end{center}}
\caption{Quantitative assessment of three synthetic depth maps in absolute difference values (per pixel) and accuracy, as reconstructed by our different models, and by LCT \cite{otoole2018}. Red regions indicate false positives and false negatives. Our three trained models miss less geometry (red regions inside contour) than LCT, mostly evidenced in the SynthBunny and SynthMannequin cases. Regarding overestimated geometry (red regions outside contour) our models tend to perform worse than LCT, being Redwood the worst case. Absolute difference values for true predicted geometry are better in LCT, but the trained models perform within acceptable margins.}
\label{fig:synth_eval_errors}
\end{figure}

\Figure{fig:synth_eval_errors} and Table~\ref{tbl:synth_error} show the quantitative assessment of the test cases above. We observe that ShapeNet and FlatNet outperform the state-of-the-art method in terms of accuracy, but exhibit a higher RMS depth error. Our method trained using the Redwood dataset tends to overestimate the existing geometry, which results in lower accuracies compared to the other methods in our test cases. This highlights the strength and weakness of our method, which is to favor shape accuracy at the expense of true depth value estimation, while LCT shows the opposite behavior. Also, our method seems to overestimate geometry pixels (outer red regions) compared to LCT, with Redwood having the highest rate of false positives. In particular, we observe that among the trained models ShapeNet exhibits the most satisfactory performance across the three test cases, as it is able to reconstruct diverse shapes with acceptable accuracy and RMS error.

\subsection{Evaluation on real-world experimental transient images}
\begin{figure}[t]
	\centering
	\rotatebox{90}{\textcolor{white}{\small Gp}} %
	\begin{minipage}[c]{.26\columnwidth}
	\centering \small \textcolor{white}{\small q}Statue\textcolor{white}{\small q}
	\end{minipage}
	\begin{minipage}[c]{.26\columnwidth}
	\centering \small \textcolor{white}{\small q}Bike\textcolor{white}{\small q}
	\end{minipage}
	\begin{minipage}[c]{.26\columnwidth}
	\centering \small \textcolor{white}{\small q}DiffuseS\textcolor{white}{\small q}
	\end{minipage}\\
	\rotatebox{90}{~\,\small\/Photo of target}
	\includegraphics[width=.26\columnwidth]{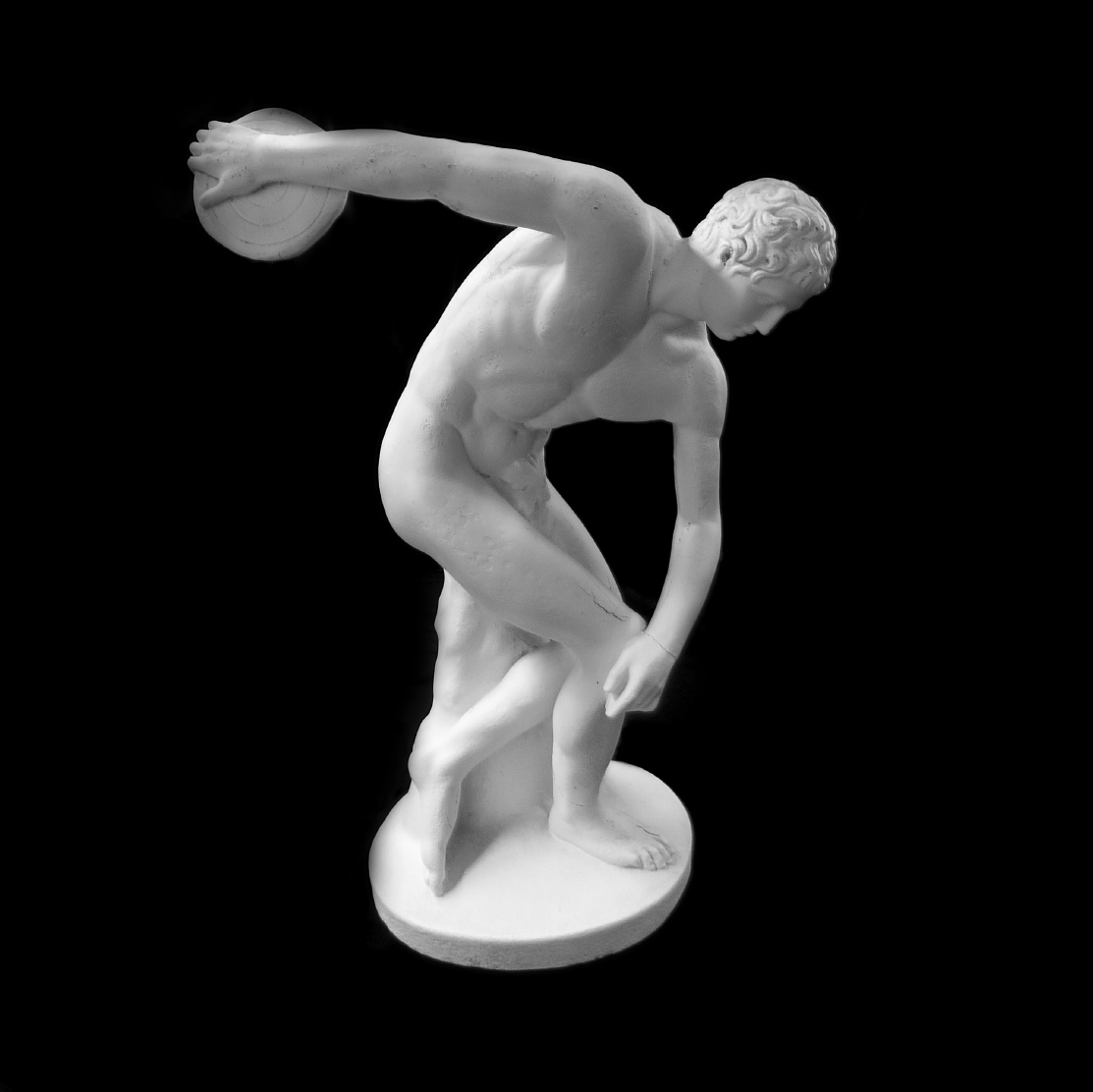}
	\includegraphics[width=.26\columnwidth]{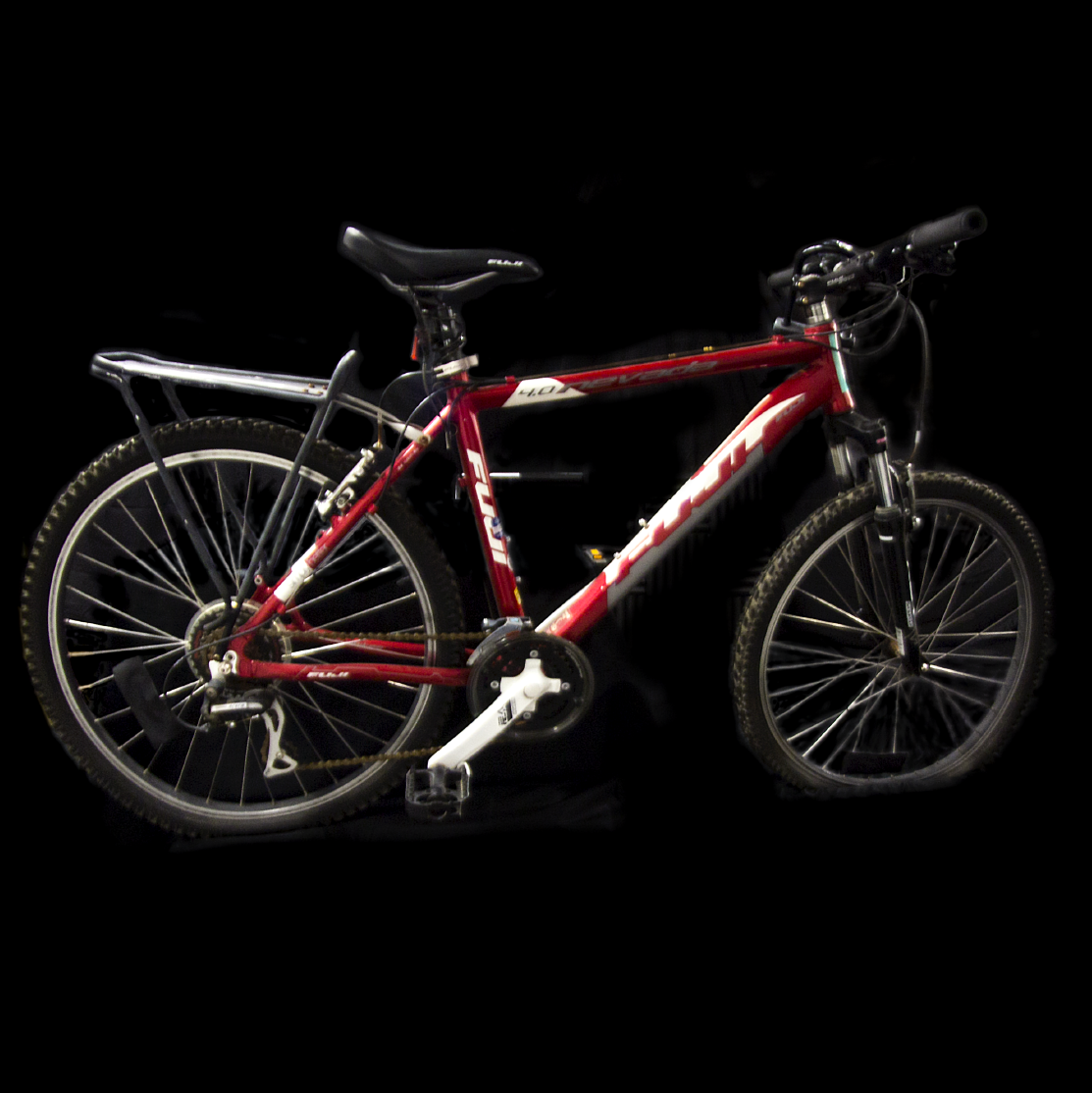}
	\includegraphics[width=.26\columnwidth]{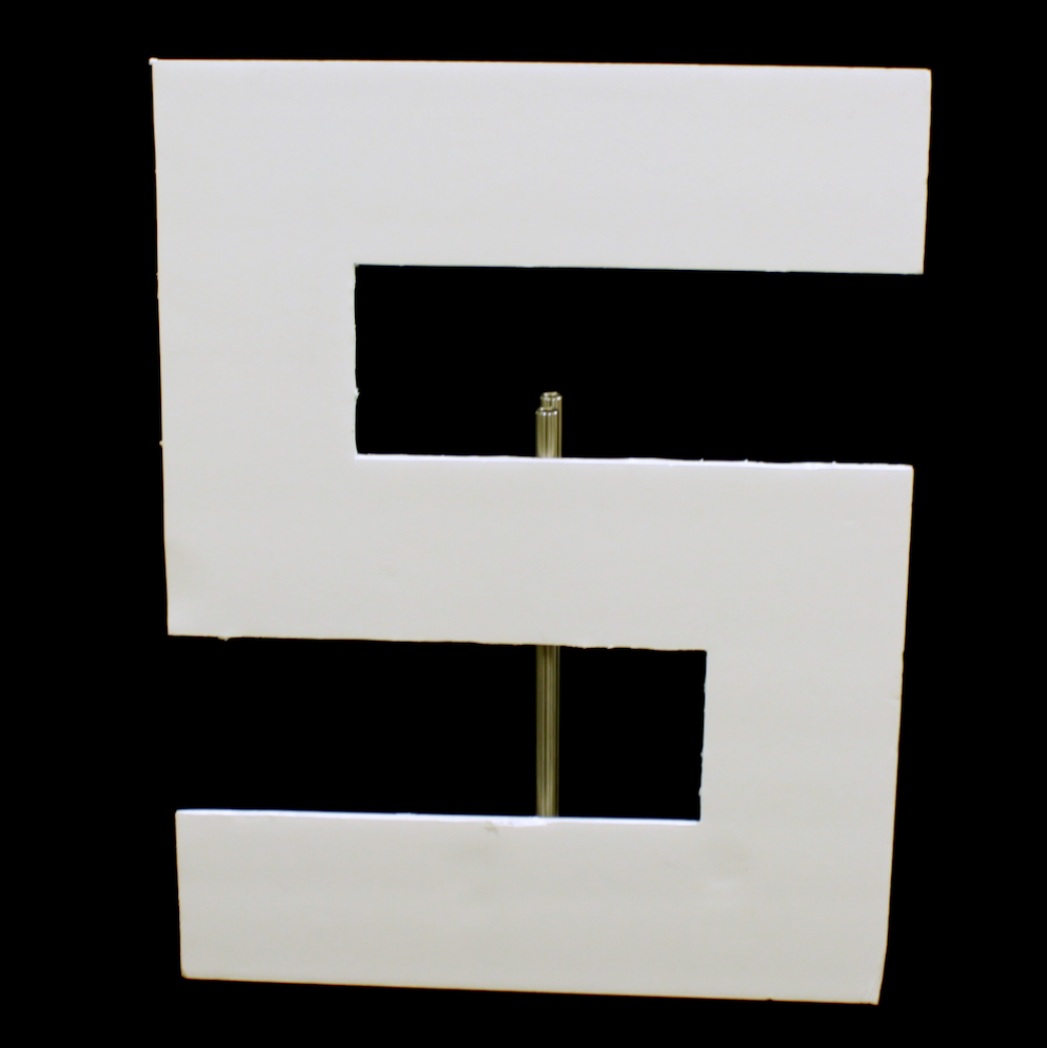}\\[0.2mm]
	\rotatebox{90}{~~~~~\small\/ShapeNet}
	\includegraphics[width=.26\columnwidth]{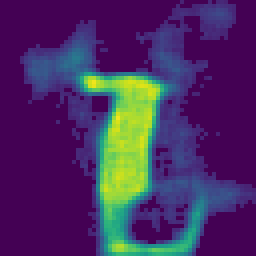}
	\includegraphics[width=.26\columnwidth]{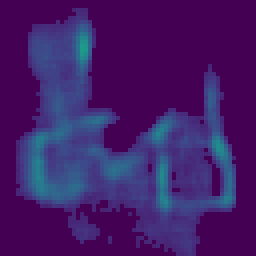}
	\includegraphics[width=.26\columnwidth]{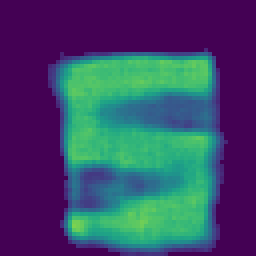}\\[0.2mm]
	\rotatebox{90}{~~~~~~~~~\small\/LCT\textcolor{white}{g}}
	\includegraphics[width=.26\columnwidth]{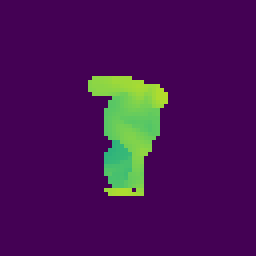}
	\includegraphics[width=.26\columnwidth]{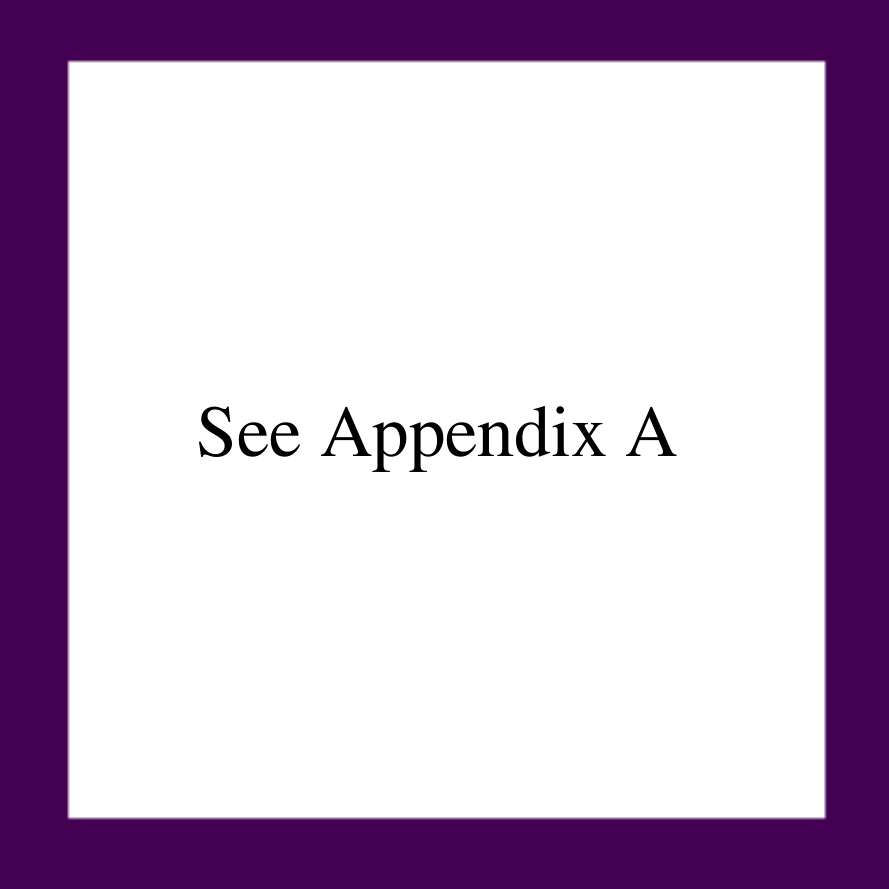}
	\includegraphics[width=.26\columnwidth]{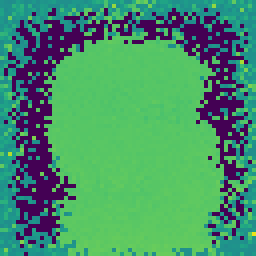}
	\caption{Predictions of three \emph{diffuse} scenes by O'Toole et al.~\cite{otoole2018} and Lindell et al.~\cite{Lindell:2019:Wave} (top) using ShapeNet (middle row) and LCT (bottom row). Lindell data has been downsampled to $64\times64$ for comparability. As a unique background is set in our method, the trained model reconstructs targets in a parameter-fashion. This allows for depth map segmentation without ambiguity. On the other hand, although LCT is able to better reconstruct complex targets (see \emph{Statue}), extracting depth maps from the resulting volumes for the \emph{Bike} and \emph{DiffuseS} can be hard (see Appendix~\ref{sec:appendix_a} for further explanation).}
	\label{fig:realworld_diffuse_results}
\end{figure}

\begin{figure}
\centering
	\rotatebox{90}{\textcolor{white}{\small Gp}} %
	\begin{minipage}[c]{.26\columnwidth}
	\centering \small \textcolor{white}{\small q}OutdoorS\textcolor{white}{\small q}
	\end{minipage}
	\begin{minipage}[c]{.26\columnwidth}
	\centering \small \textcolor{white}{\small q}SU\textcolor{white}{\small q}
	\end{minipage}
	\begin{minipage}[c]{.26\columnwidth}
	\centering \small Mannequin
	\end{minipage}\\

	\rotatebox{90}{~\,\small\/Photo of target}
	\includegraphics[width=.26\columnwidth]{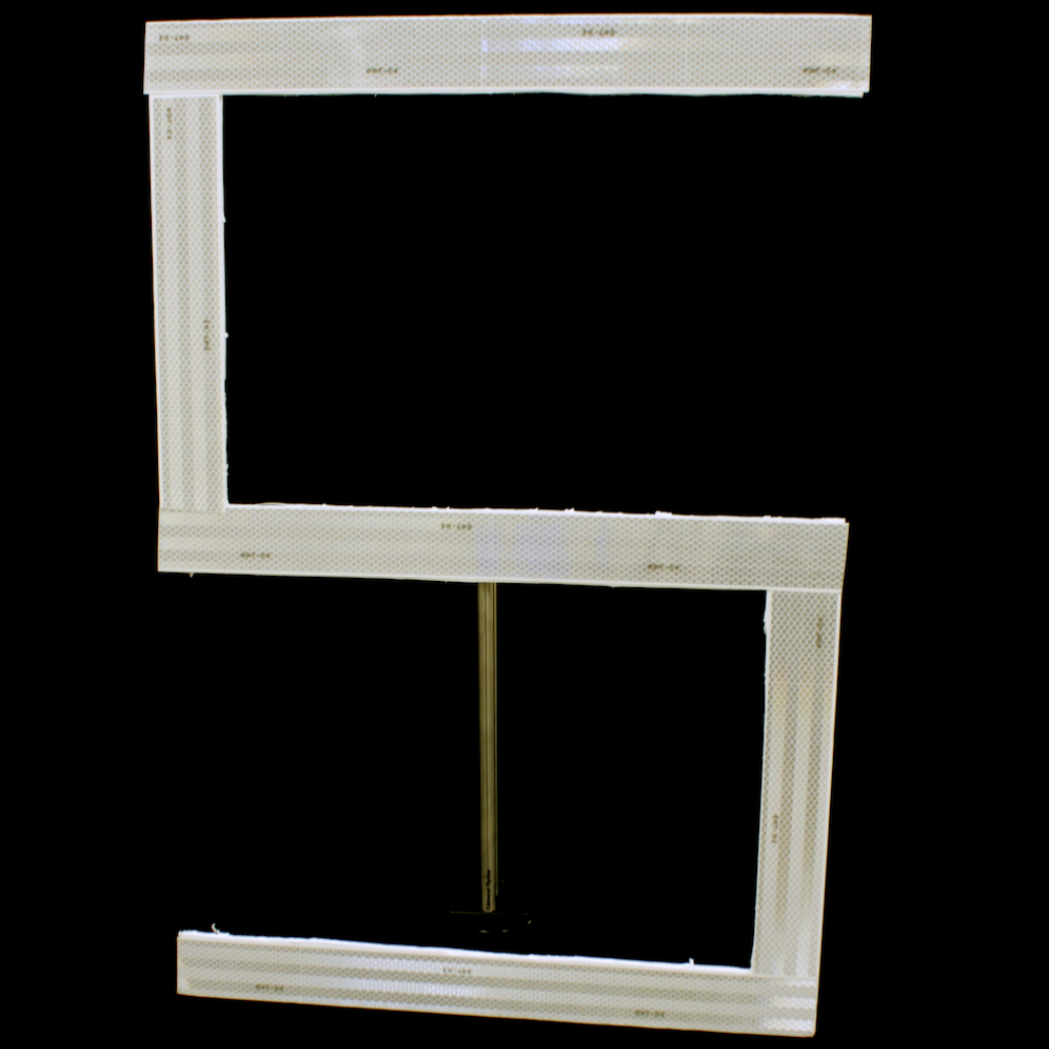}
	\includegraphics[width=.26\columnwidth]{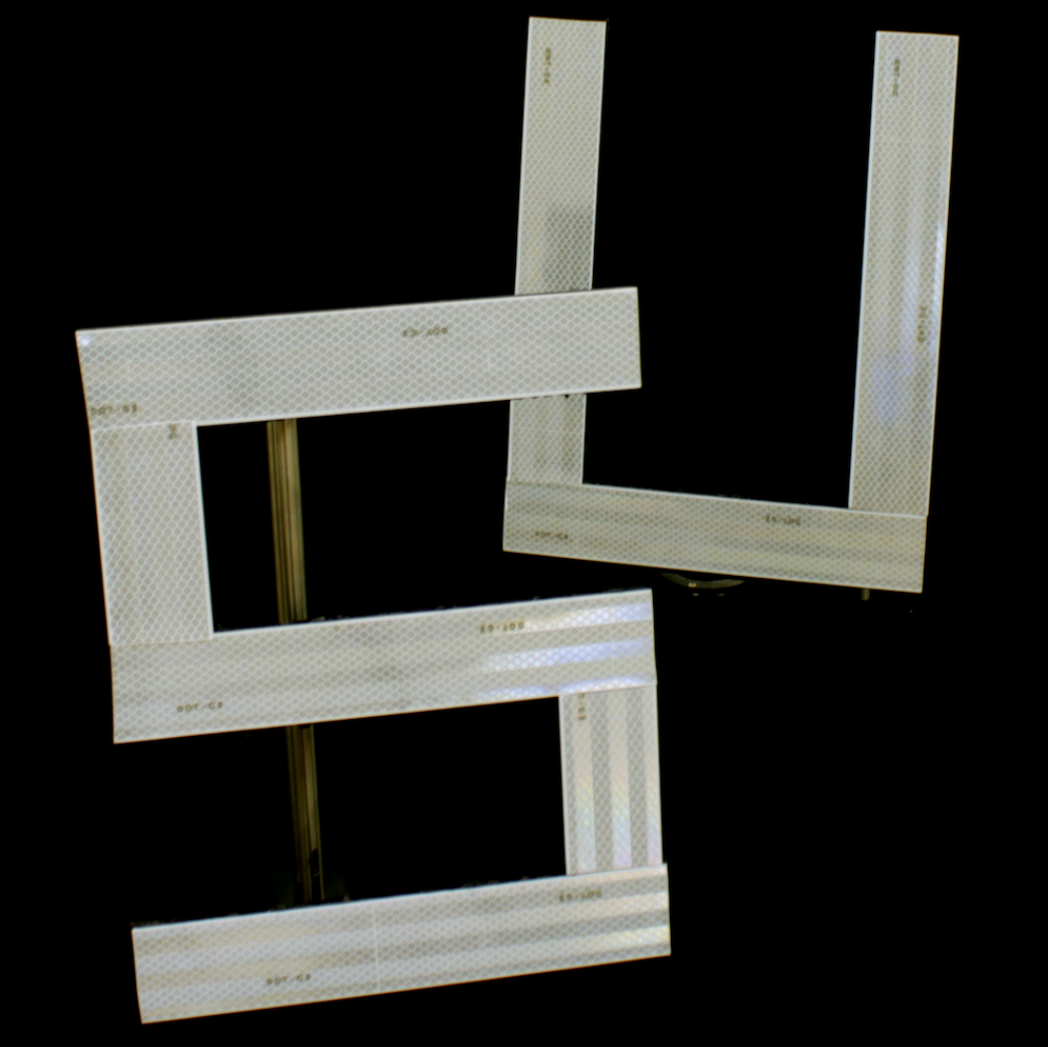}
	\includegraphics[width=.26\columnwidth]{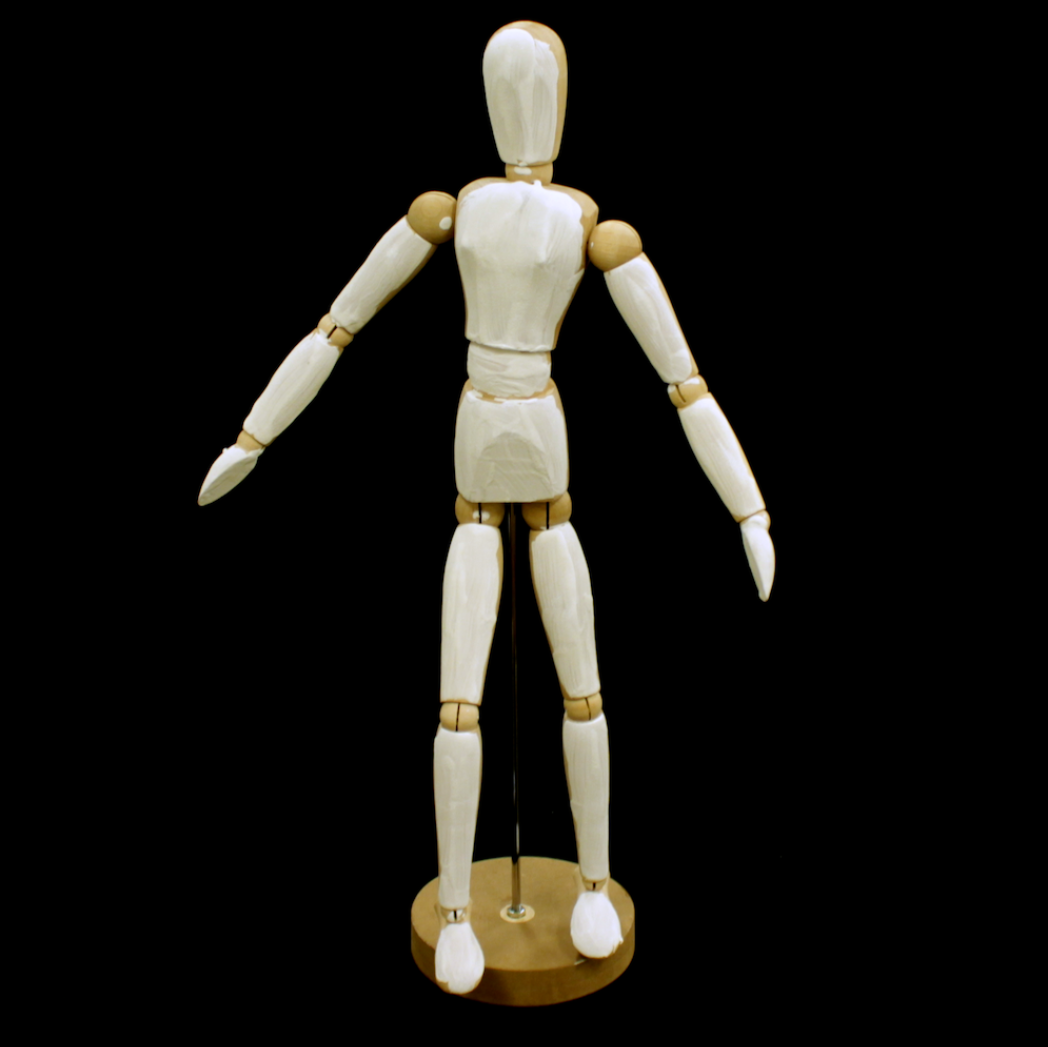}\\[0.2mm]
  \rotatebox{90}{~~~~~\small\/ShapeNet}
	\includegraphics[width=.26\columnwidth]{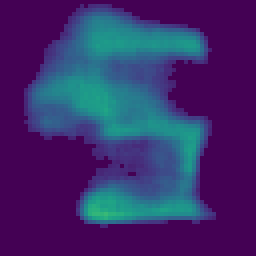}
	\includegraphics[width=.26\columnwidth]{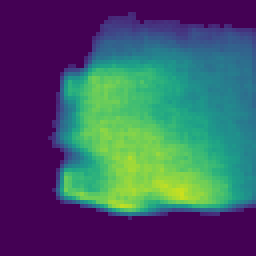}
	\includegraphics[width=.26\columnwidth]{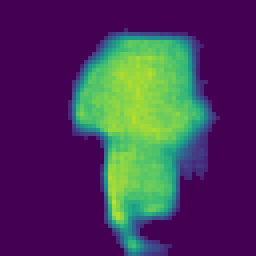}\\[0.2mm]
		\rotatebox{90}{~~~~~~~~~\small\/LCT\textcolor{white}{g}}
	\includegraphics[width=.26\columnwidth]{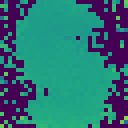}
	\includegraphics[width=.26\columnwidth]{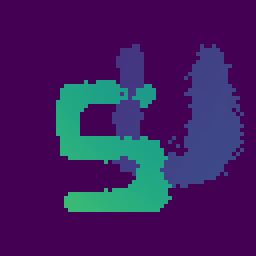}
	\includegraphics[width=.26\columnwidth]{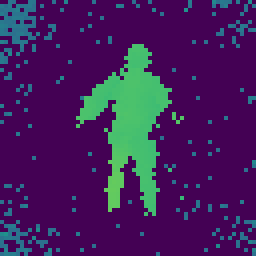}
	\caption{Predictions of three \emph{retroreflective} scenes by O'Toole et al.~\cite{otoole2018} (top) using ShapeNet (middle row) and LCT (bottom row). Note that our network has never been exposed to retroreflection during training. Thus, any retroreflective target constitutes a failure case of the presented method. Yet, our model is able to reconstruct the \emph{OutdoorS} target (left) enough to make it more recognizable than LCT. For the \emph{Mannequin} case, although far too coarse compared to the LCT prediction, the network still captures global features of the general shape.}
\label{fig:realworld_retroreflective_results}
\end{figure}

Having trained and validated our models on synthetic data, we now test the performance on inputs measured in the real world. We use data provided by O'Toole et al.~\cite{otoole2018} and Lindell et al.~\cite{Lindell:2019:Wave}, which we downsample to 32$\times$32$\times$256 and feed into our neural models. Given the overall performance outlined in the previous section, we use the ShapeNet-trained model to compute our depth maps. As baseline for our comparison, we use the LCT method, performed on the original (full-resolution) datasets. The results of this experiment are shown in Figures~\ref{fig:realworld_diffuse_results} and~\ref{fig:realworld_retroreflective_results}. Since our data augmentation does not model all intricacies of the real measurement setups, the predicted depth maps are blurrier than for the synthetic datasets. Nevertheless, most of the resulting shapes remain recognizable. Remarkably, this even holds, up to a certain degree, for the case of retroreflective targets, although our network was trained on purely diffuse inputs. Overall, such performance underlines generalization capabilities beyond assumptions considered in our forward model.

\subsection{Sensor model evaluation}\label{spad_evaluation}
\begin{figure}[t]
	\centering
	\rotatebox{90}{\textcolor{white}{\small Gp}} %
	\begin{minipage}[c]{.23\columnwidth}
	\centering \small SynthBunny
	\end{minipage}
	\begin{minipage}[c]{.23\columnwidth}
	\centering \small SynthDiff.S
	\end{minipage}
	\begin{minipage}[c]{.23\columnwidth}
	\centering \small SynthMann.
	\end{minipage}
		\begin{minipage}[c]{.23\columnwidth}
	\centering \small \textcolor{white}{\small y}DiffuseS\textcolor{white}{\small y}
	\end{minipage}\\
	\rotatebox{90}{\small~~~~~~~~Plain\textcolor{white}p}
	\includegraphics[width = 0.23\columnwidth]{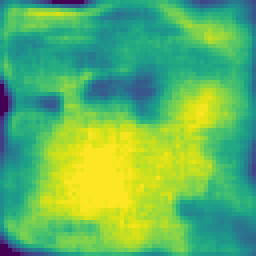}
	\includegraphics[width =  0.23\columnwidth]{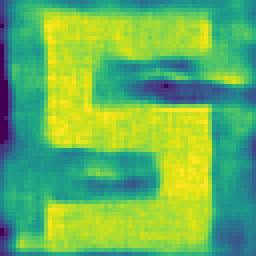}
	\includegraphics[width =  0.23\columnwidth]{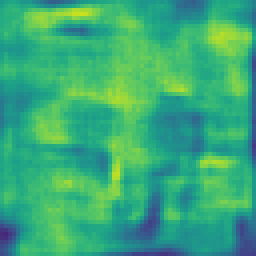}
	\includegraphics[width=.23\columnwidth]{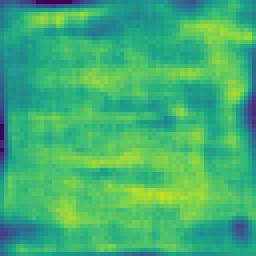}\\[0.4mm]
	\rotatebox{90}{\small~\,Poisson only\textcolor{white}p}
	\includegraphics[width =  0.23\columnwidth]{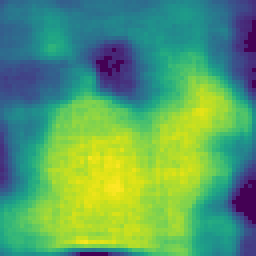}
	\includegraphics[width =  0.23\columnwidth]{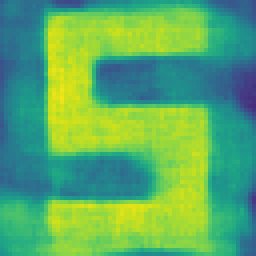}
	\includegraphics[width =  0.23\columnwidth]{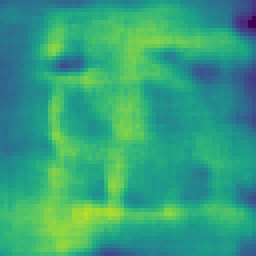}
	\includegraphics[width=.23\columnwidth]{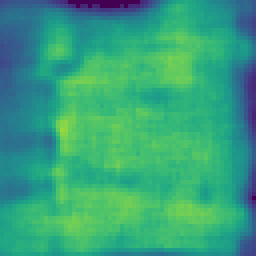}\\[0.4mm]
	\rotatebox{90}{\small\/~~~~~~~~\,Full\textcolor{white}p}
	\includegraphics[width = 0.23\columnwidth]{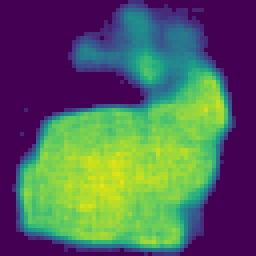}
	\includegraphics[width =  0.23\columnwidth]{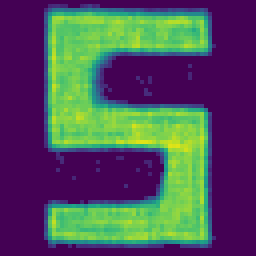}
	\includegraphics[width =  0.23\columnwidth]{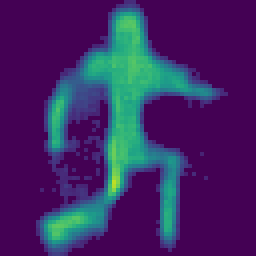}
	\includegraphics[width=.23\columnwidth]{figures/results_and_evaluation/augmented_vs_non_augmented/2-Fully_augmented_ShapeNet/real_data_otoole_S/000000.png}\\[-0.1mm]
	\caption{Adding sensor features to the forward model improves performance on noisy synthetic as well as real-world input. Only the full model with Poisson and bias recovers a recognizable depiction of the \emph{DiffuseS} target.%
	}
	\label{fig:non-aug_vs_aug_synth}
\end{figure}

We now seek to evaluate the contribution of the SPAD model augmentations for generalization to synthetic and measured data. To this end, we trained two additional versions of the ShapeNet model that include no augmentation at all (``plain'') or only noise (``Poisson''):
\begin{eqnarray}
I_{plain}(x,y,t) &=& c \cdot I(x,y,t),\label{eq:plain} \\
I_{Poisson}(x,y,t) &=& \mathcal{P}(c \cdot I(x,y,t)),\label{eq:only_poisson}
\end{eqnarray}

respectively. \Figure{fig:non-aug_vs_aug_synth} shows the impact of our sensor model on noise-degraded synthetic samples as well as on real-world input. While all models are able to predict visually recognizable renditions of \emph{SynthDiffuseS}, they perform poorly for more complicated geometries \emph{SynthBunny} and \emph{SynthMannequin}. By combining both Poisson noise and bias augmentations (\Equation{eq:spad_model}), we observe that the reconstruction quality significantly improves. On the real-world dataset \emph{DiffuseS} \cite{otoole2018}, the difference turns out even more severe. We note that more advanced sensor models might further improve the reconstruction quality further but would likely also increase the training time.

\subsection{Ablation studies}
We evaluate the performance of our pipeline in two adverse scenarios that are of high practical relevance for NLoS reconstruction: low light and low input resolution.

\begin{figure}[t]
	\centering
	\begin{minipage}[c]{.23\columnwidth}
	\includegraphics[width=\columnwidth]{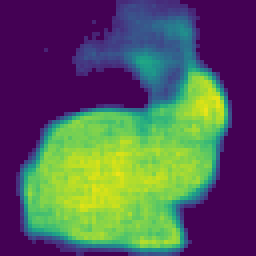}\\[-5.7em]
	\textcolor{white}{\small~28.93\,dB}\\[3.5em]
	\end{minipage}
		\begin{minipage}[c]{.23\columnwidth}
	\includegraphics[width=\columnwidth]{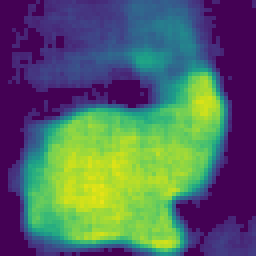}\\[-5.7em]
	\textcolor{white}{\small~17.66\,dB}\\[3.5em]
	\end{minipage}
		\begin{minipage}[c]{.23\columnwidth}
	\includegraphics[width=\columnwidth]{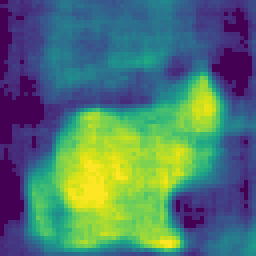}\\[-5.7em]
	\textcolor{white}{\small~10.44\,dB}\\[3.5em]
	\end{minipage}
		\begin{minipage}[c]{.23\columnwidth}
	\includegraphics[width=\columnwidth]{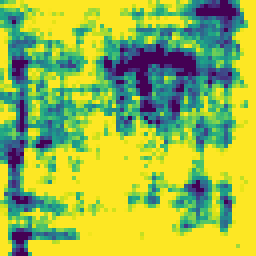}\\[-5.7em]
	\textcolor{black}{\small~--10.58\,dB}\\[3.5em]
	\end{minipage}
	\\[-0.1mm]
	\begin{minipage}[c]{.23\columnwidth}
	\includegraphics[width=\columnwidth]{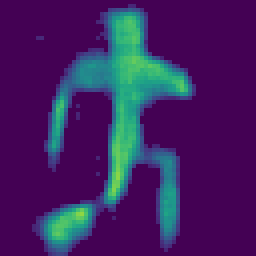}\\[-5.7em]
	\textcolor{white}{\small~14.38\,dB}\\[3.5em]
	\end{minipage}
		\begin{minipage}[c]{.23\columnwidth}
	\includegraphics[width=\columnwidth]{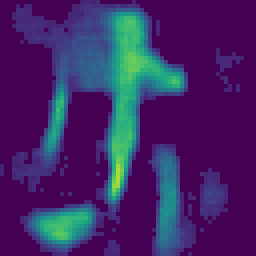}\\[-5.7em]
	\textcolor{white}{\small~12.01\,dB}\\[3.5em]
	\end{minipage}
		\begin{minipage}[c]{.23\columnwidth}
	\includegraphics[width=\columnwidth]{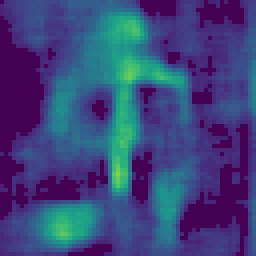}\\[-5.7em]
	\textcolor{white}{\small~4.15\,dB}\\[3.5em]
	\end{minipage}
		\begin{minipage}[c]{.23\columnwidth}
	\includegraphics[width=\columnwidth]{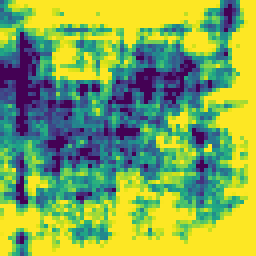}\\[-5.7em]
	\textcolor{black}{\small~--18.52\,dB}\\[3.5em]
	\end{minipage}
	\\[-0.1mm]
	\begin{minipage}[c]{.23\columnwidth}
	\includegraphics[width=\columnwidth]{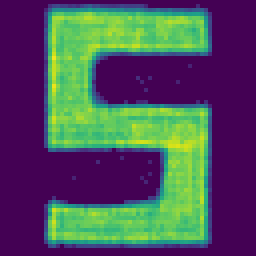}\\[-5.7em]
	\textcolor{white}{\small~30.60\,dB}\\[3.5em]
	\end{minipage}
		\begin{minipage}[c]{.23\columnwidth}
	\includegraphics[width=\columnwidth]{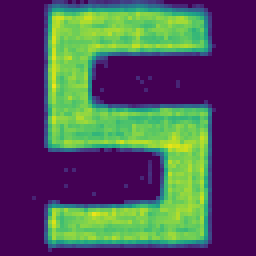}\\[-5.7em]
	\textcolor{white}{\small~23.29\,dB}\\[3.5em]
	\end{minipage}
		\begin{minipage}[c]{.23\columnwidth}
	\includegraphics[width=\columnwidth]{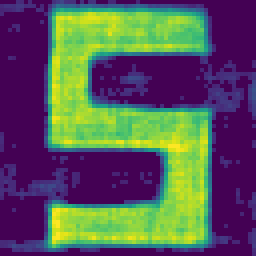}\\[-5.7em]
	\textcolor{white}{\small~13.39\,dB}\\[3.5em]
	\end{minipage}
		\begin{minipage}[c]{.23\columnwidth}
	\includegraphics[width=\columnwidth]{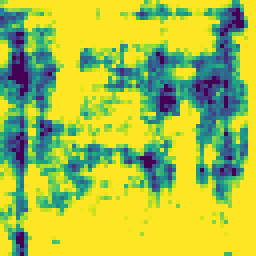}\\[-5.7em]
	\textcolor{black}{\small~--7.00\,dB}\\[3.5em]
	\end{minipage}\\[0.1em]
	\caption{Ablation study: high noise levels. Predictions for input data with high amounts of added Poisson noise. The numbers indicate peak signal-to-noise ratio of the input. Note that, even for extreme levels of noise, our network is still able to retrieve relevant features of the targets.}
	\label{poisson_series}
\end{figure}

To assess robustness under low-light conditions (light paths involving three diffuse bounces tend to be extremely lossy), we increase the level of Poisson noise on our three synthetic test cases, and feed them to the ShapeNet-trained model (\Figure{poisson_series}). The predictions are stable even under severe degradation, and only break down for extreme levels of noise.

\begin{figure}[t]
	\centering
	\begin{minipage}[c]{.03\columnwidth}
	\rotatebox{90}{\small ~SynthMannequin}
	\end{minipage}
	\begin{minipage}[c]{.23\columnwidth}
	\raggedleft
	\footnotesize Spatial\textcolor{white}{~$\nearrow$\,}\\
	 degradation $\nearrow$\,\\
	\includegraphics[width=\columnwidth]{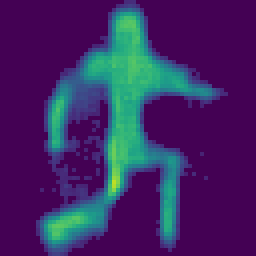}\\
		Temporal $\searrow$\,\\
	degradation\textcolor{white}{~$\searrow$\,}
	\end{minipage}%
	\hfill
\begin{minipage}[c]{.23\columnwidth}
\centering\small
	16$\times$16$\times$256\\
	\includegraphics[width=\columnwidth]{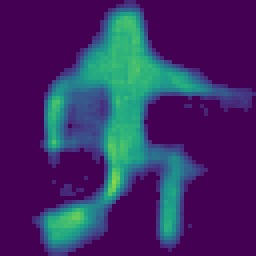}\\[0.5mm]
	\includegraphics[width=\columnwidth]{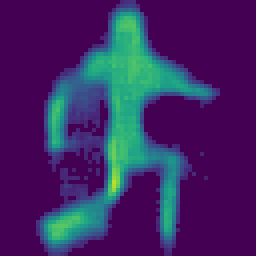}\\
	32$\times$32$\times$128
	\end{minipage}%
	\hfill
\begin{minipage}[c]{.23\columnwidth}
\centering\small
	8$\times$8$\times$256\\
	\includegraphics[width=\columnwidth]{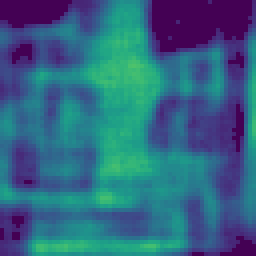}\\[0.5mm]
	\includegraphics[width=\columnwidth]{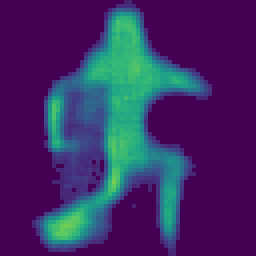}\\
	32$\times$32$\times$64
	\end{minipage}%
	\hfill
\begin{minipage}[c]{.23\columnwidth}
\centering\small
	4$\times$4$\times$256\\
	\includegraphics[width=\columnwidth]{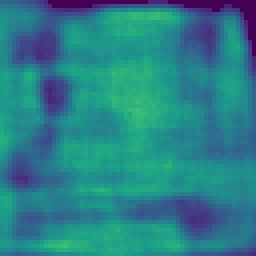}\\[0.5mm]
	\includegraphics[width=\columnwidth]{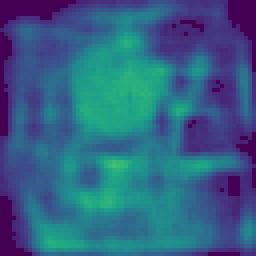}\\
	32$\times$32$\times$32
	\end{minipage}\\[2mm]%
	\centering
	\begin{minipage}[c]{.03\columnwidth}
	\rotatebox{90}{\small ~DiffuseS}
	\end{minipage}
	\hfill
	\begin{minipage}[c]{.23\columnwidth}
	\raggedleft
	\footnotesize Spatial\textcolor{white}{~$\nearrow$\,}\\
	 degradation $\nearrow$\,\\
	\includegraphics[width=\columnwidth]{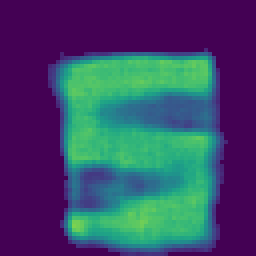}\\
		Temporal $\searrow$\,\\
	degradation\textcolor{white}{~$\searrow$\,}
	\end{minipage}%
	\hfill
		\begin{minipage}[c]{.23\columnwidth}
			\centering\small
			16$\times$16$\times$256\\
			\includegraphics[width=\columnwidth]{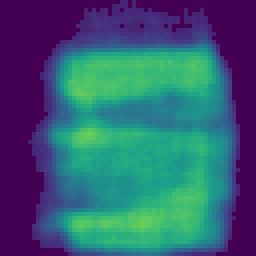}\\[0.4mm]
			\includegraphics[width=\columnwidth]{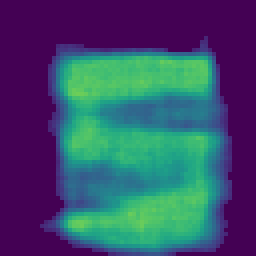}\\
			32$\times$32$\times$128
			\end{minipage}%
			\hfill
			\begin{minipage}[c]{.23\columnwidth}
				\centering\small
				8$\times$8$\times$256\\
				\includegraphics[width=\columnwidth]{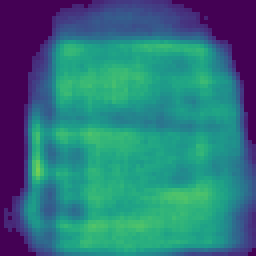}\\[0.4mm]
				\includegraphics[width=\columnwidth]{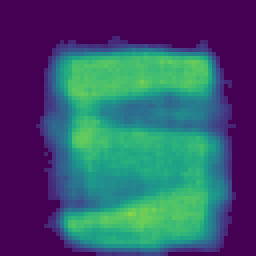}\\
				32$\times$32$\times$64
				\end{minipage}%
				\hfill
				\begin{minipage}[c]{.23\columnwidth}
					\centering\small
					4$\times$4$\times$256\\
					\includegraphics[width=\columnwidth]{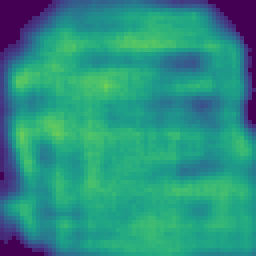}\\[0.4mm]
					\includegraphics[width=\columnwidth]{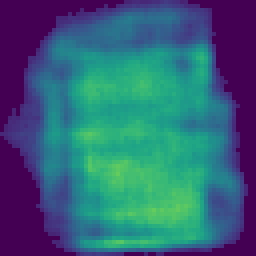}\\
					32$\times$32$\times$32
					\end{minipage}\\[2mm]%
	\caption{Ablation study: input resolution. ShapeNet predictions on spatially and temporally downsampled data. Shown are a synthetic (top) and a real-world (bottom) example. In both cases, synthetic and real measurement, the trained model is robust downsampling in spatial and temporal dimensions separately. Full resolution (far left) is 32$\times$32$\times$256.}	\label{fig:downsample_synth_real}
\end{figure}

Another interesting aspect of any reconstruction scheme is to determine the critical input resolution that is required to recover a given target. We simulate this situation by blockwise averaging of values in the input data. This is not equivalent to training a suitably dimensioned network at native resolution. Yet, it allows us to use models trained in the full resolution to perform predictions on decimated input. We independently decimated the spatial and temporal dimensions by factors of 2, 4 and 8. \Figure{fig:downsample_synth_real} shows the behavior for a synthetic and a real test case. We note that one level of spatial reduction (factor 2$\times$2) leads to acceptable results, whereas even a fourfold reduction of the temporal resolution yields virtually indistinguishable results for both the synthetic and the real test case.

\subsection{Higher-resolution reconstructions}
\begin{figure}
	\centering
	\includegraphics[width=.32\columnwidth]{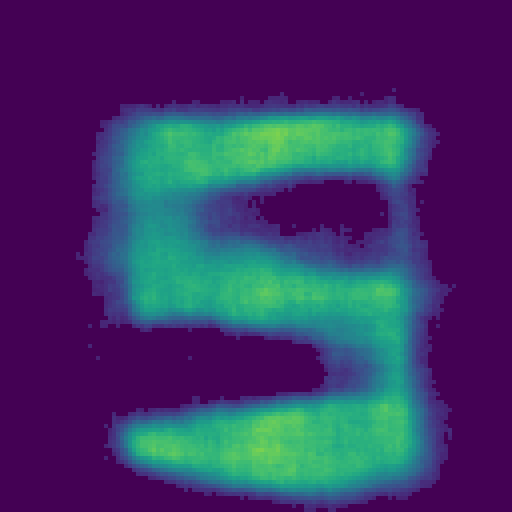}
	\includegraphics[width=.32\columnwidth]{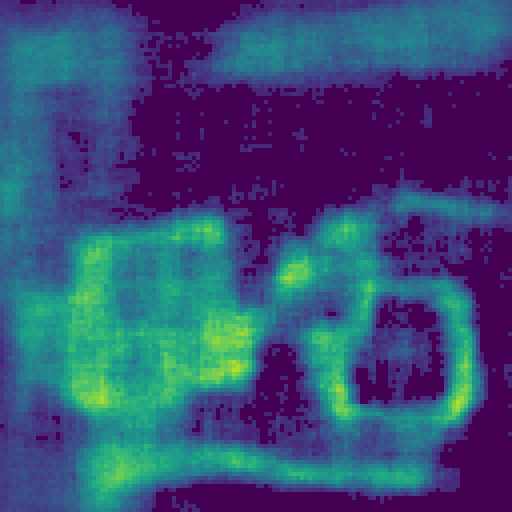}
	\includegraphics[width=.32\columnwidth]{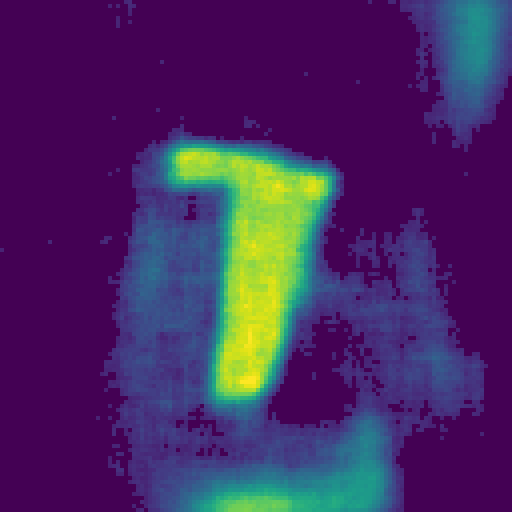}\\[0.5mm]
	\includegraphics[width=.32\columnwidth]{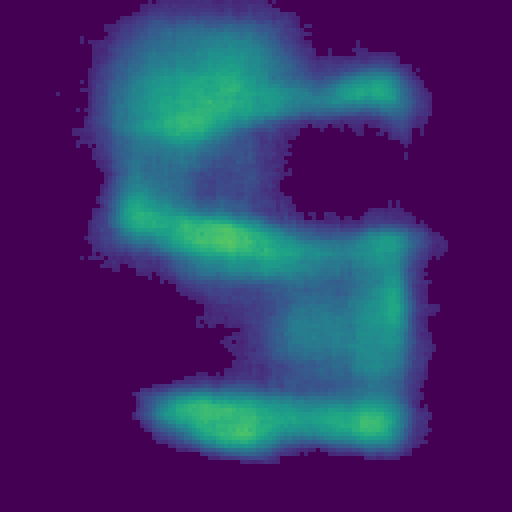}
	\includegraphics[width=.32\columnwidth]{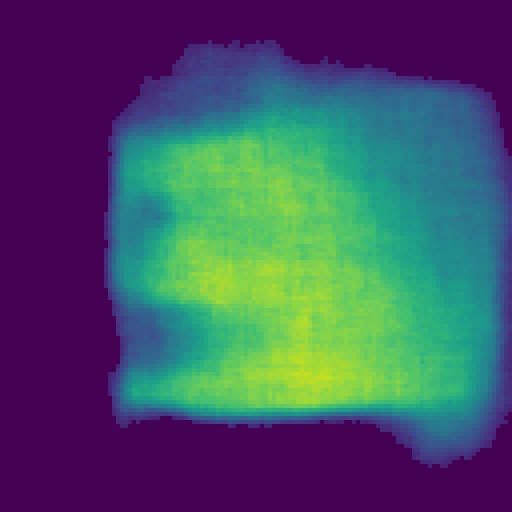}
	\includegraphics[width=.32\columnwidth]{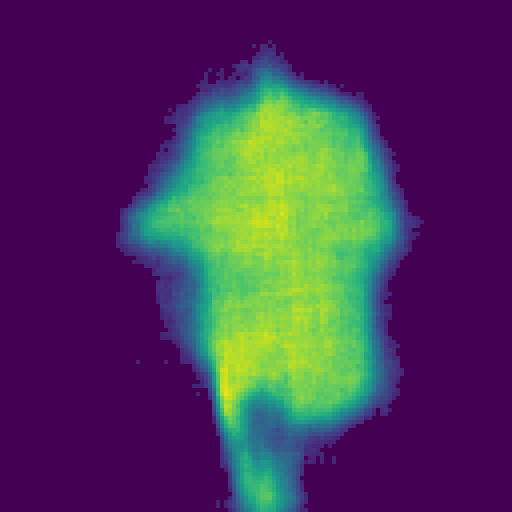}

	\caption{Multiple ShapeNet 128$\times$128 predictions on real measurements. Top: Diffuse targets. Bottom: Retroreflective targets. In the diffuse case, all targets have benefited from the higher resolution prediction as the general shapes have become shaper. For the retroreflective case, none of the targets show any improvement for this resolution}\label{high_resolution}
\end{figure}

We also explored the question of whether we can train models for higher resolution targets with the proposed pipeline. Like in the previous case, we stack layers until the output size is matched (two, for the case 128$\times$128). As fully-connected layers grow quickly with the output size, using multiple fully-connected regressor layers becomes increasingly expensive. This constitutes an inherent limitation of our architecture. For the specific case of 128$\times$128 outputs, we can still use one fully-connected layer, which, according to preliminary experiments, retains good quality. We trained on ShapeNet depth maps of 128$\times$128 while keeping the input resolution unchanged (32$\times$32$\times$256). \Figure{high_resolution} shows 128$\times$128 predictions performed on the real measurements for diffuse and retroreflective targets. By qualitatively inspecting these results, we observe that diffuse predictions have benefitted from these choices, which is not clear in the retroreflective cases.

\section{Discussion}
In this study we show that deep learning is a promising approach for non-line-of-sight geometry reconstruction.
Thanks to our sensor model, our network generalizes to real-world scenes, even though it is trained on purely synthetic data.
We are also able to show that our approach performs well for input data with extreme amounts of shot noise.
Thanks to its fast runtime, our approach might prove a competitive option for time-critical applications.
In future work, we would like to extend our scene representation to three dimensions in order to be able to account for self-occlusion in the scene.
Furthermore, it would be interesting to investigate on how to combine multiple measurements using a deep neural network. 
\small
\bibliographystyle{ieee_fullname}

\appendix
\section{Extracting depth maps from volume data}
\label{sec:appendix_a}
\begin{figure}%
	\centering
	\rotatebox{90}{\small~~~{Z-Y}%
	}~\includegraphics[width=0.92\columnwidth]{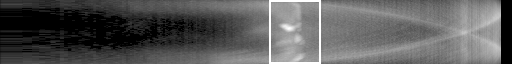}\\%
	\rotatebox{90}{\small~~~30\% Thres~~~MaxIntens\,X-Y}~\includegraphics[width=0.23\columnwidth]{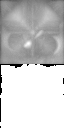}%
	\includegraphics[width=0.23\columnwidth]{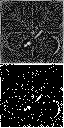}%
	\includegraphics[width=0.23\columnwidth]{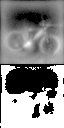}%
	\includegraphics[width=0.23\columnwidth]{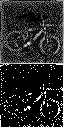}\\
	{\small~~~~~~~~~~~~~~~lct~~~~~~~~~~~~~~~~~lct\_filt~~~~~~~~~lct\_window~~~lct\_window\_filt}\\
	(a)\\[2mm]
	\rotatebox{90}{\small~~~{Z-Y}%
	}~\includegraphics[width=0.92\columnwidth]{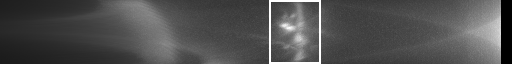}\\%
	\rotatebox{90}{\small~~~30\% Thres~~~MaxIntens\,X-Y}~\includegraphics[width=0.23\columnwidth]{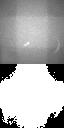}%
	\includegraphics[width=0.23\columnwidth]{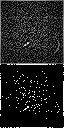}%
	\includegraphics[width=0.23\columnwidth]{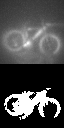}%
	\includegraphics[width=0.23\columnwidth]{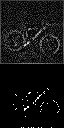}\\
	{\small~~~~~~~~~~~~~~fk~~~~~~~~~~~~~~~~~~fk\_filt~~~~~~~~~fk\_window~~~~fk\_window\_filt}\\
	(b)
	\caption{Reconstruction of a downsampled ($64\times 64\times 512$) version of the Bike30 dataset \cite{Lindell:2019:Wave} using LCT (a) and f-k migration (b), shown in max-intensity projection as proposed by the authors of the respective works. 
		At the top of each group is the reconstructed volume projected into the $z-y$ plane. Only a small portion of the volume (highlighted by a box) contains the target object.
		From the full volume and a temporally windowed version (suffix ``\_window''), we have obtained maximum-intensity projections (middle row) and attempted segmenting the object (bottom row) by setting a threshold of 30\% on the projected image or its Laplacian-filtered version (suffix ``\_filt'').}%
	\label{fig:bike}%
\end{figure}

One important observation we made on the provided material is that it can be hard to extract meaningful depth maps from volumetric solutions. In Figure~\ref{fig:bike}, we show as an example the ``Bike30'' dataset \cite{Lindell:2019:Wave} in $64\times 64$ resolution, as reconstructed using the LCT method by O'Toole et al.~\cite{otoole2018} and f-k migration by Lindell et al.~\cite{Lindell:2019:Wave} (code provided by authors).
The detailed depictions shown in the respective publications have often been cropped to a tight temporal window, without which the solution would be barely visible (Fig.~\ref{fig:bike}).

In our experience, the quality of a depth map of this type of scene mainly hinges on proper foreground-background segmentation.
Note that we are including this information not to argue about either method's performance or (dis-)advantages, but merely to illustrate the difficulties in comparing volumetric and depth map-based solutions.
 \end{document}